\documentclass[reprint,aps,superscriptaddress]{revtex4-1}
\usepackage{amsmath}
\usepackage{amssymb}
\usepackage{dcolumn}
\usepackage{graphicx}
\usepackage{url, subfigure, fancyhdr, hyperref,relsize,multirow}

\renewcommand{\Re}{\mathbb{R}}

\newcommand{\bra}[1]{\langle #1 |}
\newcommand{\ket}[1]{|#1\rangle}

\newcommand{\Tr}{\mbox{Tr}}
\newcommand{\Id}{\mathbb{I}}
\newcommand{\Ze}{\mathbb{Z}}

\newcounter{proofcounter}
\newenvironment{ProofEnv}[2]
{
  \vskip 0.1in
  \refstepcounter{proofcounter}
  \noindent {\bf #1 \arabic{proofcounter}#2:}
}
{
\hfill$\Box$
\vskip 0.2in
}

\newcounter{defncounter}
\newenvironment{DefnEnv}[1]
{
  \vskip 0.1in
  \refstepcounter{defncounter}
  \noindent {\bf Definition \arabic{defncounter} (#1):}
}
{
\hfill\break
\hspace*{\fill}\line(1,0){50}
\vskip 0.1in
}

\begin{document}

\preprint{APS/123-QED}

\title{Quantum POMDPs}

\author{Jennifer Barry}
\affiliation{Rethink Robotics}
\email{jbarry@csail.mit.edu}
\author{Daniel T. Barry}
\affiliation{Denbar Robotics}
\email{dbarry@denbarrobotics.com}
\author{Scott Aaronson}
\affiliation{MIT CSAIL}
\email{aaronson@csail.mit.edu}

\begin{abstract}
We present quantum observable Markov decision processes (QOMDPs), the quantum
analogues of partially observable Markov decision processes (POMDPs). \
In a QOMDP, an agent is acting in a world where the state is represented as a quantum state
and the agent can choose a superoperator to apply. \ This is similar to
the POMDP belief state, which is a probability distribution over world
states and evolves via a stochastic matrix. \ We show that the
existence of a policy of at least a certain value has the same
complexity for QOMDPs and POMDPs in the polynomial and infinite
horizon cases. \ However, we also prove that the existence of a policy
that can reach a goal state is decidable for goal POMDPs and undecidable
for goal QOMDPs.
\end{abstract}

\maketitle

\section{Introduction}
Partially observable Markov decision processes (POMDPs) are a
world model commonly used in artificial
intelligence~\cite{kaelbling98,pineau03,russell03,spaan05,smith04}.
POMDPs model an agent acting in a
world of discrete states. \ The world is always in exactly one state,
but the agent is not told this state.  Instead, it can
take actions and receive observations
about the world. \ The actions an agent takes are non-deterministic; before
taking an action, the
agent knows only the probability distribution of its next state given the
current state. \ Similarly, an observation does not give the agent
direct knowledge of the current world state, but the agent knows the
probability of receiving a given observation in each possible
state. \ The agent is rewarded for the actual, unknown world state
at each time step, but, although it knows the reward model, it is not
told the reward it received. \ POMDPs are often used to model robots, because
robot sensors and actuators give them a very limited understanding of
their environment.


As we will discuss further in Section~\ref{sec:pomdp}, an agent can
maximize future expected reward in a POMDP by maintaining a
probability distribution, known as a belief state, over the world's current
state. \ By carefully
updating this belief state after every action and observation, the agent can ensure
that its belief state reflects the correct probability that the world is in each
possible state. \ The agent can make decisions
using only its belief about the state without ever needing to
reason more directly about the actual world state.

In this paper, we introduce and study ``quantum observable Markov decision
processes'' (QOMDPs). \ A QOMDP is similar in spirit to a POMDP, but allows
the belief state to be a quantum state (superposition or mixed state)
rather than a simple probability distribution. \ We represent the
action and observation process jointly as a superoperator. \ POMDPs are then just
the special case of QOMDPs where the quantum state is always diagonal in some fixed basis.

Although QOMDPs are the quantum analogue of POMDPs, they have
different computability properties. \ Our main result, in this paper, is that there exists a decision problem (namely, goal state reachability) that is computable for POMDPs but uncomputable for QOMDPs.

One motivation for studying QOMDPs is simply that they're the natural quantum generalizations of POMDPs, which
are central objects of study in AI. \ Moreover, as we show here, QOMDPs have {\em different} computability properties than
POMDPs, so the generalization is not an empty one. \ Beyond this
conceptual motivation, though, QOMDPs might also find applications in
quantum control and quantum fault-tolerance. \ For example, the
general problem of controlling a noisy quantum system, given a
discrete ``library'' of noisy gates and measurements, in order to
manipulate the system to a desired end state, can be formulated as a
QOMDP. \ Indeed, the very fact that POMDPs have turned out to be such
a useful abstraction for modeling {\em classical} robots, suggests
that QOMDPs would likewise be useful for modeling control systems that
operate at the quantum scale. \ At any rate, this seems like
sufficient reason to investigate the complexity and computability
properties of QOMDPs, yet we know of no previous work in that
direction. \ This paper represents a first step.

Let us mention that soon after an earlier version of this paper was submitted here and posted on arXiv \cite{barry}
we were provided a manuscript by another group engaged in simultaneous work, Ying and Ying \cite{ying}.  They
considered quantum Markov decision processes (MDPs), and proved
undecidability results for them that are very closely related to our
results. \ In particular, these authors show that the finite-horizon
reachability problem for quantum MDPs is undecidable, and they also do
so via a reduction from the matrix mortality problem. \ Ying and Ying
also prove $\mathsf{EXP}$-hardness and
uncomputability for the infinite-horizon case (depending on whether
one is interested in reachability with probability $1$ or with
probability $p<1$, respectively). \ On the other hand, they give an
algorithm that decides, given a quantum MDP and an invariant subspace
$B$, whether or not there exists a policy that reaches $B$ with
probability $1$ regardless of the initial state; and they prove
several other results about invariant subspaces in MDPs. \ These
results nicely extend and complement ours as well as previous work by the same group \cite{ying01}

One possible advantage of the present work is that, rather than considering
(fully-observable) MDPs, we consider POMDPs. \ The latter seem to us
like a more natural starting point than MDPs for a quantum treatment,
because there is never ``full observability'' in quantum mechanics. \
Many results, including the undecidability results mentioned above,
can be translated between the MDP and POMDP settings, by the simple
expedient of considering `memoryful' MDP policies: that is, policies
that remember the initial state, as well as all actions performed so
far and all measurement outcomes obtained. \ Such knowledge is
tantamount to knowing the system's \emph{current} quantum state $\rho$.
\ However, because we consider POMDPs, which by definition can take
actions that depend on $\rho$, we never even need to deal with the
issue of memory. \ A second advantage of this work is that we
explicitly compare the quantum against the classical case (something
not done in \cite{ying}), showing why the same problem is undecidable
in the former case but decidable in the latter

Finally, we mention that there has been other work that sought to
model quantum agents in dynamic and uncertain environments
\cite{combes,tiersch}, though without formal computability and
uncomputability results.

\section{Partially Observable Markov Decision Processes (POMDPs)}
\label{sec:pomdp}

For completeness, in this section we give an overview of Markov
decision processes and POMDPs.

\subsection{Fully Observable Case}

We begin by defining fully observable Markov decision processes
(MDPs). \ This will
facilitate our discussion of POMDPs because POMDPs can be reduced to
continuous-state MDPs. \ For more details, see
Russell and Norvig, Chapter 17~\cite{russell03}.

A Markov Decision Process (MDP) is a model of an agent acting in an
uncertain but observable world. \ An MDP is a tuple $\left\langle S, A, T,
R, \gamma\right\rangle$
consisting of a set of states $S$, a set of actions $A$, a
state transition function $T(s_i, a, s_j):S\times A\times S \rightarrow
[0,1]$ giving the probability that taking action $a$ in state $s_i$
results in state $s_j$, a reward function $R(s_i, a): S\times A
\rightarrow \Re$ giving the reward of taking action $a$ in state $s_i$,
and a discount factor $\gamma\in [0,1)$ that discounts the importance
of reward gained later in time.
At each time step, the world is in exactly one, known state, and the
agent chooses
to take a single action, which transitions the world to a new state according to
$T$. \ The objective is for the agent to act in such a way as to maximize future
expected reward.

The solution to an MDP is a policy. \ A {\em policy} $\pi(s_i, t): S
\times \Ze^+\rightarrow A$ is a function mapping
states at time $t$ to actions. \ The {\em value} of a policy at state
$s_i$ over horizon $h$ is the future  expected reward of acting
according to $\pi$ for $h$ time steps:
\begin{equation}
V_\pi(s_i, h) =
\begin{array}{l}
R(s_i, \pi(s_i, h)) + \\
\gamma\sum_{s_j\in S}T(s_i,\pi(s_i, h),s_j)V_\pi(s_j, h-1)
\end{array}.
\end{equation}
The {\em solution} to an MDP of horizon $h$ is the {\em optimal policy} that maximizes future
expected reward over horizon $h$. \ The associated decision problem is
the policy existence problem:

\begin{DefnEnv}{Policy Existence Problem}
\label{defn:pep}
The {\em policy existence problem} is to decide, given a decision
process $D$, a starting state $s$, horizon $h$, and value $V$, whether
there is a policy of
horizon $h$ that achieves value at least $V$ for $s$ in $D$.
\end{DefnEnv}

For MDPs, we will evaluate the infinite horizon case. \ In this case,
we will drop the time argument from the policy since it does not
matter; the optimal policy at time infinity is the same as the optimal policy
at time infinity minus one. \ The optimal policy over an infinite
horizon is the one inducing the value function
\begin{equation}
V^*(s_i) = \max_{a\in A}\left[R(s_i, a) + \gamma\sum_{s_j\in S}T(s_i,
  a, s_j)V^*(s_j)\right].
\label{eqn:bellman}
\end{equation}
Equation~\ref{eqn:bellman} is called the {\em Bellman equation}, and there is a
unique solution for $V^*$~\cite{russell03}. \ Note that
$V^*$ is non-infinite if $\gamma < 1$. \ When the input size is
polynomial in $|S|$ and $|A|$, finding an $\epsilon$-optimal policy for an MDP
can be done in polynomial time ~\cite{russell03}.


A derivative of the MDP of interest to us is the
          {\em goal MDP}. \ A goal
MDP is a tuple $M = \left\langle S, A, T, g\right\rangle$ where $S$,
$A$, and $T$
are as before and $g \in S$ is an absorbing goal state so $T(g, a, g)
= 1$ for all $a\in A$. \ The
objective in a goal MDP is to find the policy that reaches the goal
with the highest probability. \ The associated decision problem is the
Goal-State Reachability Problem:

\begin{DefnEnv}{Goal-State Reachability Problem for Decision Processes}
\label{defn:grp}
The {\em goal-state reachability problem} is to
decide, given a goal decision process $D$ and starting state $s$,
whether there exists a policy that
can reach the goal state from $s$ in a \emph{finite} number of
steps with probability $1$.
\end{DefnEnv}

When solving goal decision processes, we never need to consider
time-dependent policies because nothing changes with the passing of
time. \ Therefore, when analyzing the goal-state reachability problem,
we will only
consider {\em stationary policies} that depend solely upon the current
state.

\subsection{Partially Observable Case}

A partially observable Markov decision process (POMDP) generalizes an
MDP to the case where the world is not fully observable. \ We follow the
work of Kaelbling et al.~\cite{kaelbling98} in explaining POMDPs.

In a
partially observable world, the agent does not know the state of the
world but
receives information about it in the form of observations. \ Formally,
a POMDP is a tuple
$\langle S, A, \Omega, T, R, O, \vec{b}_0, \gamma\rangle$
where $S$ is a set of
states, $A$ is a set of actions, $\Omega$ is a set of observations,
$T(s_i, a, s_j):S\times A\times S \rightarrow [0,1]$ is the probability
of transitioning to state $s_j$ given that action $a$ was taken in
state $s_i$, $R(s_i, a): S\times A\rightarrow \Re$ is the reward for
taking action $a$ in state $s_i$, $O(s_j, a, o):S\times A\times \Omega
\rightarrow [0,1]$ is the probability of making observation $o$ given
that action $a$ was taken and ended in state $s_j$, $\vec{b}_0$ is a
probability distribution over possible initial states, and $\gamma \in [0,
1)$ is the discount factor.

In a POMDP the world state is
``hidden'', meaning that the agent does not know the world state, but the
dynamics of the world behave according to the actual underlying state. \ At each
time step, the agent chooses an
action, the world transitions to a new state according to its
current, hidden state and $T$, and the agent receives an observation
according to the world state after the transition and $O$. \
As with
MDPs, the goal is to maximize future expected reward.

POMDPs induce a {\em belief
  MDP}. \ A {\em belief state} $\vec{b}$ is a probability distribution over
possible world states. \ For $s_i\in S$, $\vec{b}_i$ is the probability that the
world is in state $s_i$. \ Since $\vec{b}$ is a probability distribution, $0
\leq \vec{b}_i \leq 1$ and $\sum_i\vec{b}_i = 1$. \ If the agent has
belief state $\vec{b}$, takes action $a$, and receives observation $o$ the
agent's new belief state is
\begin{equation}
\begin{array}{rcl}
\vec{b'}_i &=& \Pr(s_i|o, a, \vec{b})\\
&=& \frac{\Pr\left(o|s_i,a,\vec{b}\right)\Pr\left(s_i|a,
  \vec{b}\right)}{\Pr\left(o|a,\vec{b}\right)}\\
&=&\frac{O(s_i, a, o)\sum_jT(s_j, a, s_i)\vec{b}_j}{\Pr\left(o|a,
  \vec{b}\right)}.
\end{array}
\label{eqn:bupdate}
\end{equation}
This is the belief update equation. \ $\Pr(o | a,
\vec{b} ) = \sum_kO(s_k, a,
o)\sum_jT(s_j, a, s_k)\vec{b}_j$ is independent of
$i$ and usually just computed afterwards as a normalizing factor that
causes $\vec{b}'$ to sum to $1$. \ We define the matrix
\begin{equation}
\left(\tau^{ao}\right)_{ij} = O(s_i, a, o)T(s_j, a, s_i).
\label{eqn:transmat}
\end{equation}
The belief update for seeing observation $o$ after taking action $a$
is
\begin{equation}
\vec{b'} = \frac{\tau^{ao}\vec{b}}{\left|\tau^{ao}\vec{b}\right|_1}
\label{eqn:bmupdate}
\end{equation}
where $\left|\vec{v}\right|_1 = \sum_i\vec{v}_i$ is the $L_1$-norm. \
The probability of transitioning
from belief state $\vec{b}$ to belief state $\vec{b'}$ when taking
action $a$ is
\begin{equation}
\tau(\vec{b}, a, \vec{b'}) = \sum_{o\in\Omega}
\Pr(\vec{b}' | a, \vec{b}, o)
\Pr(o | a, \vec{b} )
\label{eqn:tau}
\end{equation}
where
$$\Pr(\vec{b'} | a, \vec{b}, o ) = \left\{\begin{array}{ll}
1 & \mbox{if } \vec{b'} =
\frac{\tau^{ao}\vec{b}}{\left|\tau^{ao}\vec{b}\right|_1}\\
0 & \mbox{else}.
\end{array}\right.$$
The expected reward of taking action $a$ in belief state $\vec{b}$ is
\begin{equation}
r(\vec{b}, a) = \sum_i\vec{b}_iR(s_i, a).
\label{eqn:rho}
\end{equation}

Now the agent always knows its belief state so the belief space is
fully observable. \ This means we can define the {\em belief MDP} $\left\langle
B, A, \tau, r, \gamma\right\rangle$ where $B$ is the set of
all possible belief states. \ The
optimal solution to the MDP is also the optimal solution to the
POMDP. \ The problem is that the state space of the belief state MDP is continuous, and all
known algorithms for solving MDPs optimally in polynomial time are
polynomial in the size of the state space. \
It was shown in 1987 that the policy existence problem for POMDPs is
$\mathsf{PSPACE}$-hard~\cite{papadimitriou87}. \ If the horizon is polynomial
in the size of the input, the policy existence problem is in
$\mathsf{PSPACE}$~\cite{kaelbling98}. \
The policy existence
problem for POMDPs in the infinite horizon case, however, is
undecidable~\cite{madani99}.

A {\em goal POMDP} is a tuple $P = \langle S,
A, \Omega, T, O, \vec{b}_0, g \rangle$
where $S$, $A$, $\Omega$, $T$, and $O$ are defined as before but
instead of a reward function, we assume that $g\in S$ is a goal
state. \ This state $g$ is absorbing so we are promised that
for all $a\in A$, that $T(g, a, g) = 1$. \ Moreover, the agent receives
an observation $o_{|\Omega|}\in \Omega$ telling it that it has reached the
goal so for all
$a\in A$, $O(g, a, o_{|\Omega|}) = 1$. \ This observation is only received in
the goal state so for all $s_i \neq g$, and all $a\in A$,
$O(s_i, a, o_{|\Omega|}) = 0$. \ The solution to a goal POMDP is a policy that
reaches the goal state with the highest possible probability starting
from $\vec{b}_0$.

We will show that because the goal is absorbing and known, the
observable belief space
corresponding to a goal POMDP is a goal MDP $M(P) = \langle B, A,
\tau, \vec{b}_0, \vec{b}_g \rangle$. \ Here $\vec{b}_g$ is the state in
which the agent knows it is in $g$ with probability $1$. \ We show that this
state is absorbing. \ Firstly the probability of observing $o$ after taking
action $a$ is
\begin{align*}
\Pr(o | a, \vec{b}_g) &= \sum_{j}O(s_j, a, o)\sum_iT(s_i, a,
s_j)(\vec{b}_g)_i\\
&= \sum_jO(s_j, a,
o)T(g, a, s_j)\\
&= O(g, a, o)\\
&= \delta_{oo_{|\Omega|}}.
\end{align*}
Therefore, if the agent has belief $\vec{b}_g$, regardless of the action
taken, the agent sees observation $o_{|\Omega|}$. \ Assume the agent takes action $a$
and sees observation $o_{|\Omega|}$. \ The next belief state is
\begin{align*}
\vec{b'}_j &= \Pr(s_j | o_{|\Omega|}, a, \vec{b}_g )\\
&= \frac{O(s_j, a,
  o_{|\Omega|})\sum_iT(s_i, a,
  s_j)\vec{b}_i}{\Pr (o_{|\Omega|} | a, \vec{b}_g )}\\
&= O(s_j, a,o_{|\Omega|})T(g, a, s_j)\\
&= \delta_{gs_j}.
\end{align*}
Therefore, regardless of the action taken, the next belief state is
$\vec{b}_g$ so this is a goal MDP.

\section{Quantum Observable Markov Decision Processes (QOMDPs)}
\label{sec:qomdp}

A quantum observable Markov decision process (QOMDP) generalizes
a POMDP by using quantum states rather than belief states. \ In a QOMDP,
an agent can apply a set of possible operations to a $d$-dimensional
quantum system. \ The operations each have ${\cal K}$ possible outcomes.
At each time step, the
agent receives an observation corresponding to the outcome of the
previous operation and can choose another operation to apply. \ The
reward the agent receives is the expected value of some operator in
the system's current quantum state.

\subsection{QOMDP Formulation}

A QOMDP uses superoperators to express both actions and
observations. \ A quantum superoperator ${\bf S} = \{K_1, ..., K_{\cal
  K}\}$ acting on
states of dimension $d$ is defined by
${\cal K}$ $d\times d$ Kraus matrices \footnote{Actually, the quantum
  operator acts on a product state of which the first dimension
  is $d$. \ In order to create quantum states of dimension $d$
  probabilistically, the superoperator entangles the possible next
  states with a measurement register and then measures that register. \
  Thus the operator actually acts on the higher-dimensional product
  space, but for the purposes of this discussion, we can treat it as
  an operator that probabilistically maps states of dimension $d$ to
  states of dimension $d$.} \cite{superoperator}. \ A set of matrices $\{K_1,
..., K_{\cal K}\}$ of dimension $d$ is a set of Kraus matrices if and only if
\begin{equation}
\sum_{i=1}^{\cal K} K_i^\dag K_i = \Id_d.
\end{equation}
If ${\bf S}$ operates on a density matrix $\rho$, there are ${\cal K}$
possible next states for $\rho$. \ Specifically the next state is
\begin{equation}
\rho'_i \rightarrow \frac{K_i \rho K_i^\dag}{\Tr(K_i \rho
  K_i^\dag)}
\label{eqn:kup}
\end{equation}
with probability
\begin{equation}
\Pr(\rho'_i | \rho) = \Tr(K_i \rho K_i^\dag).
\label{eqn:kprob}
\end{equation}
The superoperator returns observation $i$ if the $i^{\text{th}}$ Kraus matrix
was applied.

We can now define the quantum observable Markov decision process
(QOMDP).
\begin{DefnEnv}{QOMDP}
\label{defn:qomdp}
A QOMDP is a tuple $\left\langle S, \Omega, {\cal A}, {\cal
  R},
\gamma, \rho_0\right\rangle$ where
\begin{itemize}
\item $S$ is a Hilbert space. \ We allow pure and mixed quantum states
  so we will represent states in $S$ as
  density matrices.
\item $\Omega = \{o_1, ..., o_{|\Omega|}\}$ is a set of
  possible observations.
\item ${\cal A} = \left\{A^1, ..., A^{|{\cal A}|}\right\}$ is a set of
  superoperators. \ Each superoperator $A^a = \{A^a_1, ...,
  A^a_{|\Omega|}\}$ has
  $|\Omega|$ Kraus matrices. \ Note that each superoperator returns the same set
  of possible observations; if this is not true in reality, some of
  the Kraus matrices may be the all zeroes matrix. \ The return of
  $o_i$ indicates the application of the $i$th
  Kraus matrix so taking action $a$ in state $\rho$ returns
  observation $o_i$ with probability
  \begin{equation}
    \Pr(o_i | \rho, a) = \Tr\left(A^a_i\rho{A^a_i}^\dag\right).
    \label{eqn:obsprob}
  \end{equation}
  If $o_i$ is observed after taking action $a$ in state $\rho$, the
  next state is
  \begin{equation}
    N(\rho, a, o_i) =
    \frac{A_i^a\rho{A^a_i}^\dag}{\Tr\left(A^a_i\rho{A^a_i}^\dag\right)}.
      \label{eqn:stateev}
  \end{equation}
\item ${\cal R} = \{R_1, ..., R_{|\cal A|}\}$ is a set of
  operators. \ The reward associated with taking action $a$ in state
  $\rho$ is the expected value of operator $R_a$ on $\rho$,
  \begin{equation}
    R(\rho, a) = \Tr(\rho R_a).
    \label{eqn:qreward}
  \end{equation}
\item $\gamma \in [0, 1)$ is a discount factor.
\item $\rho_0 \in S$ is the starting state.
\end{itemize}
\end{DefnEnv}
Like an MDP or POMDP, a QOMDP represents a world in which an agent
chooses actions at discrete time steps and receives observations. \ The
world modeled by the QOMDP is a quantum system that begins in
$\rho_0$, the starting state of the QOMDP. \ At
each time step, the agent chooses a superoperator from the set ${\cal
  A}$, whereupon the corresponding operation is done on the system and the agent
receives an observation from the set $\Omega$ in accordance with
the laws of quantum mechanics. \ The
agent also receives a reward according to the state of the system
after the operation and ${\cal
  R}$. \ As in an MDP or POMDP, the agent knows the entire QOMDP model
a priori and its goal is to use this information to maximize its
future expected reward.

A QOMDP is fully observable in the same sense that the belief state
MDP for a POMDP is fully observable. \ Just as the agent in a POMDP
always knows its belief state, the agent in a QOMDP always knows the
current quantum superposition or mixed state of the system. \ In a
POMDP, the agent can update its belief state when it takes an action and
receives an observation using equation~\ref{eqn:bmupdate}. \ Similarly, in a
QOMDP, the agent can keep track of the quantum state using
equation~\ref{eqn:stateev} each time it takes an action and receives
an observation. \ Note that a QOMDP
is much more analogous to the belief state MDP of a POMDP than to the
POMDP itself. \ In a POMDP, the system is always in one, actual
underlying world state that is simply unknown to the agent; in a QOMDP,
the system can be in a superposition state for which no underlying
``real'' state exists.

As with MDPs, a policy for a QOMDP is a function $\pi: S\times \Ze^+
\rightarrow {\cal A}$ mapping states at time $t$ to actions. \ The
value of the policy over horizon $h$ starting from state $\rho_0$ is
$$V^\pi(\rho_0) = \sum_{t=0}^h E\left[\left.\gamma^tR(\rho_t,
    \pi(\rho_t))\right| \pi\right].$$
Let $\pi_h$ be the policy at time $h$. \ Then
\begin{multline}
V^{\pi_h}(\rho_0) = R(\rho_0, \pi_h(\rho_0)) +\\
\gamma\sum_{i =
  1}^{|\Omega|}\Pr(o_i | \rho_0,
\pi_h(\rho_0))V^{\pi_{h-1}}(N(\rho_0, \pi_h(\rho_0), o_i))
\label{eqn:vf}
\end{multline}
where $\Pr(o_i | \rho_0, \pi_h(\rho_0))$, $N(\rho_0,
\pi_h(\rho_0), 
o_i)$, and $R(\rho_0, \pi_h(\rho_0))$ are defined by
equations~\ref{eqn:obsprob}, \ref{eqn:stateev}, and~\ref{eqn:qreward}
respectively. \ The Bellman equation (equation~\ref{eqn:bellman}) still
holds using these definitions.

A {\em goal QOMDP} is a tuple $\left\langle S,
          \Omega, {\cal A}, \rho_0, \rho_g\right\rangle$ where $S$,
          $\Omega$, ${\cal A}$, and $\rho_0$ are as defined above. \
          The goal state $\rho_g$ must be absorbing so that for all
          $A^i \in {\cal A}$ and all $A^i_j \in A^i$ if
          $\Tr(A^i_j\rho_g {A^i_j}^\dag) > 0$ then
          $$\frac{A^i_j\rho_g{A^i_j}^\dag}{\Tr(A^i_j\rho_g
            {A^i_j}^\dag)} = \rho_g.$$
          As with goal MDPs and POMDPs, the objective for a goal QOMDP
          is to maximize the probability of reaching the goal state.

\subsection{QOMDP Policy Existence Complexity}

As we can always simulate classical evolution with a quantum
system,
the definition of QOMDPs contains POMDPs. \ Therefore we immediately
find that the policy
existence problem for QOMDPs in the infinite horizon case is undecidable. \ We
also find that the polynomial horizon case is
$\mathsf{PSPACE}$-hard. \ We can, in fact, prove that the polynomial horizon
case is in $\mathsf{PSPACE}$.

\begin{ProofEnv}{Theorem}{}
\label{pf:qpspace}
The policy existence problem (Definition~\ref{defn:pep}) for QOMDPs
with a polynomial horizon is in $\mathsf{PSPACE}$.

{\bf Proof:}  Papadimitriou and Tsitsiklis~\cite{papadimitriou87}
showed that polynomial horizon POMDPs are in $\mathsf{PSPACE}$ and the
proof still holds for QOMDPs with the appropriate substitution for the
calculations of the probability of an observation given a quantum
state and action [Eq.~\ref{eqn:obsprob}], $N$ [Eq.~\ref{eqn:stateev}],
and $R$ [Eq.~\ref{eqn:qreward}], all of which can clearly be done in
$\mathsf{PSPACE}$ when the horizon is polynomial.
\end{ProofEnv}

\section{A Computability Separation in Goal-State Reachability}
\label{sec:comp}

However, although the policy existence problem has the same complexity
for QOMDPs and POMDPs, we can show that the goal-state
reachability problem (Definition~\ref{defn:grp}) is
decidable for goal POMDPs but undecidable for goal QOMDPs.

\subsection{Undecidability of Goal-State Reachability for QOMDPs}

We will show that the goal-state reachability problem is undecidable
for QOMDPs by showing that we can reduce the quantum measurement
occurrence problem proposed by Eisert et al.~\cite{eisert12} to it.

\begin{DefnEnv}{Quantum Measurement Occurrence Problem}
\label{defn:qmop}
The {\em quantum measurement occurrence problem} (QMOP) is to decide, given
a quantum superoperator described by ${\cal K}$ Kraus operators
${\bf S} = \{K_1, ..., K_{\cal K}\}$, whether there is some finite sequence
$\{i_1, ..., i_n\}$ such that $K_{i_1}^\dag... K_{i_n}^\dag
K_{i_n}... K_{i_1} = 0$.
\end{DefnEnv}

\begin{figure}
\centering
\includegraphics[width=\columnwidth]{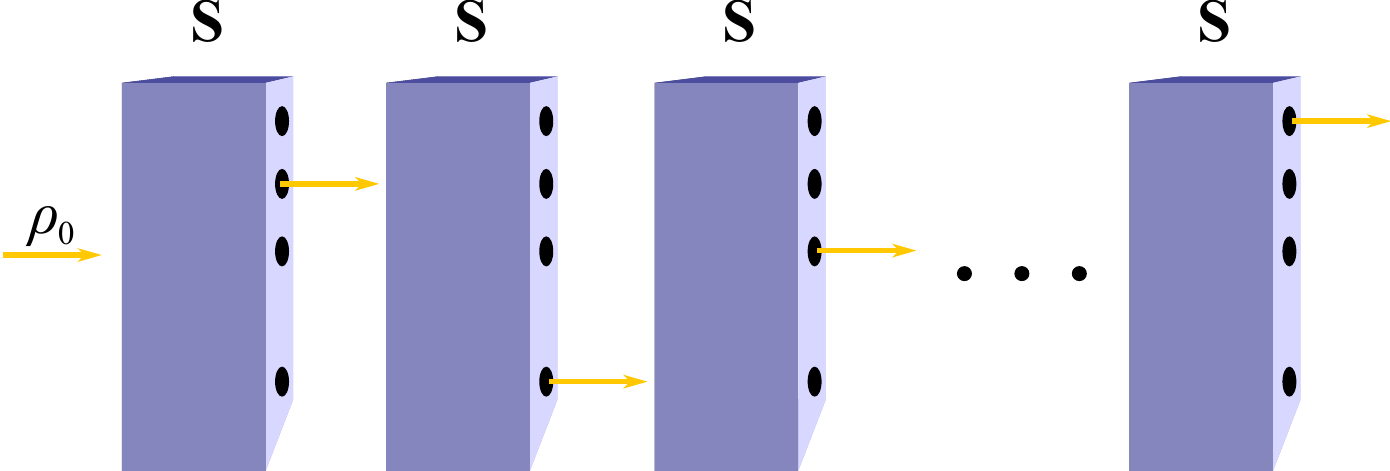}
\caption{(Color online) The quantum measurement occurrence problem. \ The starting
  state $\rho_0$ is fed into the superoperator ${\bf S}$. \ The output
  is then fed iteratively back into ${\bf S}$. \ The question is whether
  there is some finite sequence of observations that can never occur.}
\label{fig:qmop}
\end{figure}

The setting for this problem is shown in Figure~\ref{fig:qmop}. \ We
assume that the system starts in state $\rho_0$. \ This state is fed
into ${\bf S}$. \ We then take the output of ${\bf
  S}$ acting on $\rho_0$ and feed that again into ${\bf S}$ and
iterate. \ QMOP is
equivalent to
asking whether there is some finite sequence of observations $\{i_1, ...,
i_n\}$ that can never occur even if $\rho_0$ is full
rank. \ We will reduce from the version of the problem
given in Definition~\ref{defn:qmop}, but will use the language of measurement
occurrence to provide intuition.

\begin{ProofEnv}{Theorem}{ (Undecidability of QMOP)}
\label{pf:qmop}
The quantum measurement occurrence problem is undecidable.

{\bf Proof:} This can be shown using a reduction from the matrix
mortality problem. \ For the full proof see Eisert et al \cite{eisert12}.
\end{ProofEnv}

\begin{figure}[t!]
\centering
\includegraphics[width=\columnwidth]{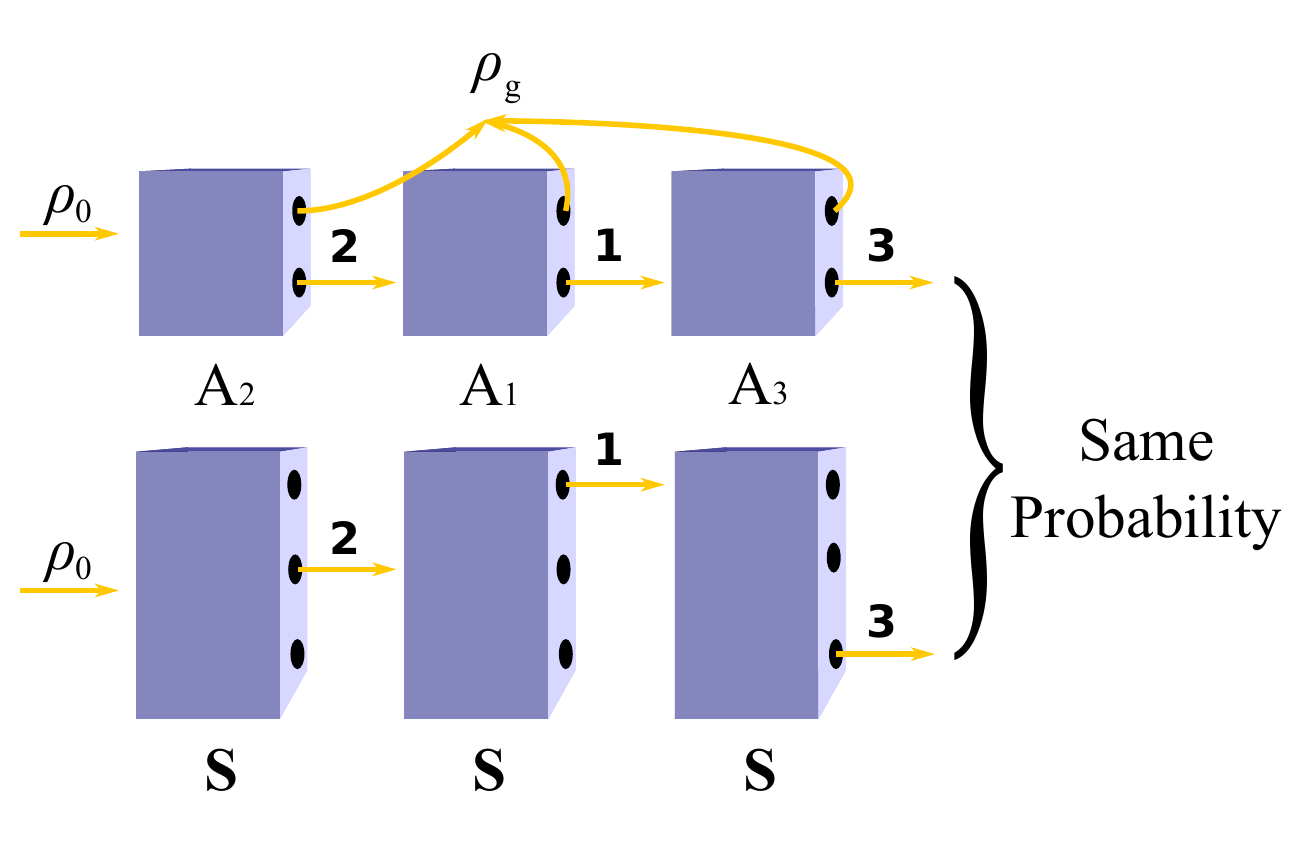}
\caption{(Color online) A goal QOMDP for a QMOP instance with superoperator ${\bf S}
  = \left\{K_1, K_2, K_3\right\}$
  with 3 possible outcomes.
  We create 3 actions to correspond to the 3 outputs of the
  superoperator. \ Each action $A_i$ has two possible outcomes: either
  the system
  transitions according to $K_i$ from ${\bf S}$ or it transitions to
  the goal state. \ Intuitively, we can think of $A_i$ as either
  outputting the observation ``transitioned to goal'' or observation
  $i$ from ${\bf S}$. \ Then it is clear that if the action
  sequence $\{A_2, A_1, A_3\}$ is taken, for instance, the probability that we
  do {\em not} see the observation sequence 2, 1, 3 is the probability
  that the system transitions to the goal state somewhere in this
  sequence. \ Therefore, the probability that an action sequence
  reaches the goal state is the probability that the corresponding
  observation sequence is not observed.
}
\label{fig:qmop_qomdp}
\end{figure}

We first describe a method for creating a goal QOMDP from an instance
of QMOP. \ The main ideas behind the choices we make here are shown in
Figure~\ref{fig:qmop_qomdp}.

\begin{DefnEnv}{QMOP Goal QOMDP}
Given an instance of QMOP with superoperator ${\bf S} = \{K_1, ...,
K_{\cal K}\}$ and
Kraus matrices of dimension $d$, we create a goal
QOMDP $Q({\bf S}) = \left\langle S, \Omega, {\cal A}, \rho_0, \rho_g\right\rangle$ as follows:
\begin{itemize}
\item $S$ is $(d+1)$-dimensional Hilbert space.
\item $\Omega = \{o_1, o_2, ..., o_{d+2}\}$ is a set of $d+2$ possible
  observations. \ Observations $o_1$ through $o_{d+1}$ correspond to
  At-Goal while $o_{d+2}$ is Not-At-Goal.
\item ${\cal A} = \left\{A^1, ..., A^{\cal K}\right\}$ is a set of ${\cal K}$
  superoperators each with $d+2$ Kraus matrices $A^i = \{A^i_1, ...,
  A^i_{d+2}\}$ each of dimension $d+1\times d+1$. \ We set
  \begin{equation}
  A^i_{d+2} = K_i \oplus 0 = \left[\begin{array}{ccc}
      \multicolumn{2}{c}{\mathlarger{\mathlarger{\mathlarger{\mathlarger{K_i}}}}}
      &\begin{array}{c}
         0\\
         \vdots
       \end{array}\\
      0 & ... & 0
    \end{array}\right],
  \end{equation}
  the $i$th Kraus matrix from the QMOP superoperator with the $d+1$st
  column and row all zeros. \
  Additionally, let
  \begin{align}
  Z^i &= \mathbb{I}_{d+1} - {A^i_{d+2}}^\dag A^i_{d+2}\\
  &= \left(\sum_{j\neq i}K_j^\dag K_j\right)\oplus 1\\
  &= \left[\begin{array}{cccc}
      \multicolumn{3}{c}{
        \mathlarger{\mathlarger{\mathlarger{
                \sum_{j\neq i}K_j^\dag K_j}}}}
      &\begin{array}{c}
         0\\
         0\\
         \vdots
       \end{array}\\
      0 & 0& ... & 1
    \end{array}\right].
  \end{align}
  Now $(K_j^\dag K_j)^\dag = K_j^\dag K_j$ and the sum of
  Hermitian matrices is Hermitian so
  $Z^i$ is Hermitian. \ Moreover, $K_j^\dag K_j$ is positive
  semidefinite, and positive semidefinite matrices are closed under positive addition, so $Z^i$ is positive semidefinite as well. \ Let an
  orthonormal eigendecomposition of $Z^i$ be
  $$Z^i = \sum_{j=1}^{d+1} z^i_j\ket{z^i_j}\bra{z^i_j}.$$
  Since $Z^i$ is a positive semidefinite Hermitian matrix, $z^i_j$ is
  nonnegative and real so $\sqrt{z^i_j}$ is also real. \
  We let $A^i_j$ for $j < d+2$ be the $d+1\times d+1$ matrix in which
  the first $d$
  rows are all 0s and the bottom row is $\sqrt{z^i_j}\bra{z^i_j}$:
  \begin{eqnarray*}
  \left({A^i_{j<d+2}}\right)_{pq} &=&
  \sqrt{z^i_j}\bra{z^i_j}q\rangle\delta_{p(d+1)},\\
  A^i_{j<d+2} &=& \left[\begin{array}{ccc}
      0 & ... & 0\\
      \vdots & \ddots & \vdots\\
      0 & ... & 0\\
      \multicolumn{3}{c}{\sqrt{z^i_j}\bra{z^i_j}}
    \end{array}\right].
  \end{eqnarray*}
  (Note that if $z^i_j = 0$ then $A^i_j$ is the all-zero matrix, but it
  is cleaner to allow each action to have the same number of Kraus
  matrices.)
\item $\rho_0$ is the maximally mixed state ${\rho_0}_{ij} =
  \frac{1}{d+1}\delta_{ij}$.
\item $\rho_g$ is the state $\ket{d+1}\bra{d+1}$.
\end{itemize}
\end{DefnEnv}

The intuition behind the definition of $Q({\bf S})$ is shown in
Figure~\ref{fig:qmop_qomdp}. \ Although each action actually has $d+2$
choices, we will show that $d+1$ of those choices (every one except
$A^i_{d+2}$) always transition to the goal state. \ Therefore action
$A^i$ really only provides two possibilities:
\begin{enumerate}
\item Transition to goal state.
\item Evolve according to $K_i$.
\end{enumerate}
Our proof will proceed as follows: Consider choosing some sequence of
actions $A^{i_1}, ...,
A^{i_n}$. \ The probability that the system transitions to the goal state is
the same as the probability that it does not evolve according to first
$K_{i_1}$ then $K_{i_2}$ etc. \ Therefore, the system transitions to the goal
state with probability $1$ if and only if it is impossible for it to
transition according
to first $K_{i_1}$ then $K_{i_2}$ etc. \ Thus in the original problem,
it must have been impossible to see the observation sequence $\{i_1,
..., i_n\}$. \ In other words, the agent can reach a goal state with
probability $1$ if and only if there is some sequence of observations in
the QMOP instance that can never occur. \ Therefore we can use
goal-state reachability in QOMDPs to solve QMOP, giving us that
goal-state reachability for QOMDPs must be undecidable.

We now formalize the sketch we just gave. \ Before we can do
anything else, we must show that $Q({\bf S})$ is in fact a goal
QOMDP. \ We start by showing that $\rho_g$ is absorbing in two lemmas.
In the first, we prove that $A^i_{j < d+2}$ transitions all density
matrices to the goal state. \ In the second, we show that $\rho_g$ has
zero probability of evolving according to $A^i_{d+2}$.

\begin{ProofEnv}{Lemma}{}
\label{pf:togoal}
Let ${\bf S} = \left\{K_1, ..., K_{\cal K}\right\}$ with Kraus matrices of
dimension $d$ be the superoperator from an instance of QMOP and let
$Q({\bf S}) = \left\langle S, \Omega, {\cal A}, \rho_0, \rho_g\right\rangle$ be the
corresponding goal QOMDP. \ For any
density matrix $\rho \in S$, if $A^i_j$ is the $j^{\text{th}}$ Kraus
matrix of the $i^{\text{th}}$ action of $Q({\bf S})$ and $j < d+2$ then
$$\frac{A^i_j\rho {A^i_j}^\dag}{\Tr(A^i_j\rho
  {A^i_j}^\dag)} = \ket{d+1}\bra{d+1}.$$

{\bf Proof:} Consider
\begin{align}
(A^i_j\rho {A^i_j}^\dag)_{pq} &= \sum_{h,
  l}{A^i_j}_{ph}\rho_{hl}{A^i_j}^\dag_{lq}\\
&= \sum_{h,l}{A^i_j}_{ph}\rho_{hl}{A^i_j}^*_{ql}\\
&= z^i_j\sum_{h,
  l}\bra{z^i_j}h\rangle\rho_{hl}\bra{l}z^i_j\rangle
\delta_{p(d+1)}\delta_{q(d+1)}
\end{align}
so only the lower right element of this matrix is nonzero. \ Thus dividing
by the trace gives
\begin{equation}
\frac{A^i_j\rho {A^i_j}^\dag}{\Tr(A^i_j \rho
  {A^i_j}^\dag)} = \ket{d+1}\bra{d+1}.
\end{equation}
\end{ProofEnv}

\begin{ProofEnv}{Lemma}{}
\label{pf:goalabs}
Let ${\bf S}$ be the superoperator from an instance of QMOP and let
$Q({\bf S}) = \{S,
\Omega, {\cal A}, \rho_0, \rho_g\}$ be the corresponding QOMDP. \ Then
$\rho_g$ is absorbing.

{\bf Proof:} By Lemma~\ref{pf:togoal}, we know that for $j < d+2$, we have
$$\frac{A^i_j\ket{d+1}\bra{d+1}{A^i_j}^\dag}{\Tr(A^i_j\ket{d+1}\bra{d+1}{A^i_j}^\dag)}
= \rho_g.$$
Here we
show that $\Tr(A^i_{d+2}\rho_g {A^i_{d+2}}^\dag) = 0$ so
that the probability of applying $A^i_{d+2}$ is $0$. \ We have:
\begin{align}
&\Tr\left(A^i_{d+2} \ket{d+1}\bra{d+1} {A^i_{d+2}}^\dag\right)\\
&=\sum_p\sum_{hl}{A^i_{d+2}}_{ph}\delta_{h(d+1)}\delta_{l(d+1)}{A^i_{d+2}}^*_{pl}\\
&= \sum_p{A^i_{d+2}}_{p(d+1)}{A^i_{d+2}}^*_{p(d+1)} = 0
\end{align}
since the $(d+1)^{\text{st}}$ column of $A^i_{d+2}$ is all zeros by
construction. \ Therefore, $\rho_g$ is absorbing.
\end{ProofEnv}

Now we are ready to show that $Q({\bf S})$ is a goal QOMDP.

\begin{ProofEnv}{Theorem}{}
\label{pf:qomdp}
Let ${\bf S} = \left\{K_1, ..., K_{\cal K}\right\}$ be the
superoperator from an instance of QMOP with
Kraus matrices of dimension $d$. \
Then $Q({\bf S}) = \left\langle S, \Omega, {\cal A}, \rho_0,
\rho_g\right\rangle$ is a goal QOMDP.

{\bf Proof:} We showed in Lemma~\ref{pf:goalabs} that $\rho_g$ is
absorbing, so all that remains to show is that the actions are
superoperators. \ Let $A^i_j$ be
the $j^{\text{th}}$ Kraus matrix of action
$A^i$. \ If $j < d+2$ then
\begin{align}
({A^i_j}^\dag {A^i_j})_{pq} &= \sum_h
{A^i_j}^\dag_{ph}{A^i_j}_{hq}\\
&= \sum_h{A^i_j}^*_{hp}{A^i_j}_{hq}\\
&= \sqrt{z^i_j}^*\bra{p}z^i_j\rangle \sqrt{z^i_j}\bra{z^i_j}q\rangle\\
&= z^i_j\bra{p}z^i_j\rangle\bra{z^i_j}q\rangle
\end{align}
where we have used that $\sqrt{z^i_j}^* = \sqrt{z^i_j}$ because
$\sqrt{z^i_j}$ is real. \
Thus for $j < d+2$
$${A^i_j}^\dag {A^i_j} = z^i_j\ket{z^i_j}\bra{z^i_j}.$$

Now
\begin{align}
\sum_{j=1}^{d+2} {A^i_j}^\dag A^i_j &= {A^i_{d+2}}^\dag A^i_{d+2} + \sum_{j=1}^{d+1} z^i_j
\ket{z^i_j}\bra{z^i_j}\\
&= {A^i_{d+2}}^\dag A^i_{d+2} + Z^i\\
&= \mathbb{I}_{d+1}.
\end{align}
Therefore $\{A^i_j\}$ is a set of Kraus matrices.
\end{ProofEnv}

Now we want to show that the probability of not reaching a goal state
after taking actions $\left\{A^{i_1},..., A^{i_n}\right\}$ is the same
as the probability of observing the sequence $\{i_1, ..., i_n\}$. \
However, before we can do that, we must take a short detour to show
that the fact that the goal-state reachability problem is defined for
state-dependent policies does not give it any advantage. \ Technically,
a policy for a QOMDP
is not time-dependent but
state-dependent. \ The QMOP problem is essentially time-dependent: we
want to know about a specific sequence of observations over time. \  A
QOMDP policy, however, is 
state-dependent: the choice of action depends not upon the number of
time steps, but upon the current state. \  When reducing a QMOP
problem to a QOMDP problem, we need to ensure that the observations
received in the QOMDP are dependent on time in the same way that they are in
the QMOP instance. \  We will be able to do this because we have designed
the QOMDP to which we reduce a QMOP instance such that
after $n$ time steps there is at most one possible non-goal
state for the system. \ The existence of such a state and the exact state that is
reachable depends upon the policy 
chosen, but regardless of the policy, there will be at most one. \ This
fact, which we will prove in the following lemma, allows 
us to consider the policy for these QOMDPs as
time-dependent: the action the time-dependent policy chooses at time step $n$ is the action
the state-dependent policy chooses for the only non-goal state the
system could possibly reach at time $n$.

\begin{ProofEnv}{Lemma}{}
\label{pf:choice}
Let ${\bf S} = \left\{K_1, ..., K_{\cal K}\right\}$ with Kraus matrices of
dimension $d$ be the superoperator from an instance of QMOP and let
$Q({\bf S}) = \left\langle S, \Omega, {\cal A}, \rho_0,
\rho_g\right\rangle$ be the
corresponding goal QOMDP. \ Let $\pi:S \rightarrow {\cal A}$
be any policy for $Q({\bf S})$. \ There is always at most one state
$\sigma_n \neq \rho_g$ such that $\Pr(\sigma_n | \pi, n) > 0$.

{\bf Proof:} We proceed by induction on $n$.

{\em Base Case $(n = 1)$}:  After $1$ time step, the agent has taken a single
action, $\pi(\rho_0)$.\  Lemma~\ref{pf:togoal} gives us that
there is only a single possible state besides $\rho_g$ after the
application of this action.

{\em Induction Step}: Let $\rho_n$ be the state on the $n^{\text{th}}$ time step
and let $\rho_{n-1}$ be the state on the $(n-1)^{\text{st}}$ time step. \ Assume
that there are only two possible choices
for $\rho_{n-1}$: $\sigma_{n-1}$ and $\rho_g$. \ If $\rho_{n-1} =
\rho_g$, then $\rho_n = \rho_g$ regardless of $\pi(\rho_g)$. \ If
$\rho_{n-1} = \sigma_{n-1}$, the agent takes action $\pi(\sigma_{n-1})
= A^{i_n}$. \ By Lemma~\ref{pf:togoal} there is only a single possible
state besides $\rho_g$ after the application of $A^{i_n}$.
\end{ProofEnv}

Thus in a goal QOMDP created from a QMOP instance, the state-dependent
policy $\pi$ can be considered a ``sequence of actions'' by looking at
the actions it will apply to each possible non-goal state in order.

\begin{DefnEnv}{Policy Path}
\label{defn:path}
Let ${\bf S} = \left\{K_1, ..., K_{\cal K}\right\}$ with Kraus matrices of
dimension $d$ be the superoperator from a QMOP instance and let
$Q({\bf S}) = \left\langle S, \Omega, {\cal A}, \rho_0, \rho_g\right\rangle$ be the
corresponding goal QOMDP. \ For any policy $\pi$ let $\sigma_k$ be the
non-goal state with nonzero probability after $k$ time steps of
following $\pi$ if it exists. \ Otherwise let $\sigma_k = \rho_g$.
Choose $\sigma_0 = \rho_0$. \ The sequence $\{\sigma_k\}$ is the {\em policy path}
for policy $\pi$. \ By Lemma~\ref{pf:choice}, this sequence is unique so
this is well-defined.
\end{DefnEnv}

We have one more technical problem we need to address before we can
look at how states evolve under policies in a goal QOMDP. \ When we
created the goal QOMDP, we added a dimension to the Hilbert space so that we
could have a defined goal state. \ We need to show that we can consider
only the upper-left $d\times d$ matrices when looking at evolution
probabilities.

\begin{ProofEnv}{Lemma}{}
\label{pf:trace}
Let ${\bf S} = \left\{K_1, ..., K_{\cal K}\right\}$ with Kraus matrices of
dimension $d$ be the superoperator from a QMOP instance and let
$Q({\bf S}) = \left\langle S, \Omega, {\cal A}, \rho_0,
\rho_g\right\rangle$ be the
corresponding goal QOMDP. \
Let $M$ be any $(d+1)\times (d+1)$ matrix and $d(M)$ be the upper left
$d\times d$ matrix in which the $(d+1)^{\text{st}}$ column and row of $M$ have
been removed. \ Then for any action $A^i \in {\cal A}$,
$$A^i_{d+2} M {A^i_{d+2}}^\dag = K_i d(M) K_i \oplus 0.$$

{\bf Proof:} We consider the multiplication element-wise:
\begin{align}
(A^i_{d+2}M{A^i_{d+2}}^\dag)_{pq} &=
\sum_{h,l=1}^{d+1}{A^i_{d+2}}_{ph}M_{hl}{A^i_{d+2}}^\dag_{lq}\\
&= \sum_{h,l=1}^d{A^i_{d+2}}_{ph}M_{hl}{A^i_{d+2}}^*_{ql}
\end{align}
where we have used that the $(d+1)^{\text{st}}$ column of $A^i_{d+2}$ is
$0$ to limit the sum. \ Additionally, if $p = d+1$ or $q = d+1$, the
sum is $0$ because the $(d+1)^{\text{st}}$ row of $A^i_{d+2}$ is $0$. \ Assume
that $p < d+1$ and $q < d+1$. \ Then
\begin{multline}
\sum_{h,l=1}^d{A^i_{d+2}}_{ph}M_{hl}{A^i_{d+2}}^*_{ql}\\
=\sum_{h,l=1}^d{K_i}_{ph}M_{hl}{K_i}_{lq}^\dag = \left(Kd(M)
  K^\dag\right)_{ql}.
\end{multline}
Thus
\begin{equation}
A^i_{d+2}M {A^i_{d+2}}^\dag = K_id(M) K_i^\dag \oplus 0.
\end{equation}
\end{ProofEnv}

We are now ready to show that any path that does not terminate in the
goal state in the goal QOMDP corresponds to some possible path through
the superoperator in the QMOP instance.

\begin{ProofEnv}{Lemma}{}
\label{pf:evol}
Let ${\bf S} = \left\{K_1, ..., K_{\cal K}\right\}$ with Kraus matrices of
dimension $d$ be the superoperator from a QMOP instance and let
$Q({\bf S}) = \left\langle S, \Omega, {\cal A}, \rho_0,
\rho_g\right\rangle$ be the
corresponding goal QOMDP. \ Let $\pi$ be any policy for $Q$ and let
$\{\sigma_k\}$ be the policy path for $\pi$. \ Assume
$\pi(\sigma_{k-1}) = A^{i_k}$. \ Then
$$\sigma_k = \frac{K_{i_k}...K_{i_1}d(\rho_0)
  K_{i_1}^\dag... K_{i_k}^\dag \oplus 0}{\Tr(K_{i_k}...K_{i_1}d(\rho_0)
  K_{i_1}^\dag... K_{i_k}^\dag)}.$$

{\bf Proof:}  We proceed by induction on $k$.

{\em Base Case $(k = 1)$}: If $k = 1$ then either some $A^{i_1}_l$
with $l < d+2$ or $A^{i_1}_{d+2}$ is applied to the system. \ In the
first case, Lemma~\ref{pf:togoal} gives us that the state becomes
$\rho_g$. \ Therefore, $\sigma_1$ is the result of applying
$A^{i_1}_{d+2}$ so
\begin{align}
\sigma_1 &= \frac{A^{i_1}_{d+2}\rho_0
  {A^{i_1}_{d+2}}^\dag}{\Tr(A^{i_1}_{d+2}\rho_0
  {A^{i_1}_{d+2}}^\dag)}\\
&= \frac{K_{i_1}d(\rho_0)K_{i_1}^\dag \oplus 0}{\Tr(K_{i_1}d(\rho_0)
  K_{i_1}^\dag\oplus 0)}\label{eqn:utr}\\
&= \frac{K_{i_1}d(\rho_0)
  K_{i_1}^\dag \oplus 0}{\Tr(K_{i_1}d(\rho_0)
  K_{i_1}^\dag)}\label{eqn:trz}
\end{align}
using Lemma~\ref{pf:trace} for Equation~\ref{eqn:utr} and the fact
that $\Tr(A\oplus 0) = \Tr(A)$ for Equation~\ref{eqn:trz}.

{\em Induction Step:}  On time step $k$, we have $\rho_{k-1} =
\sigma_{k-1}$ or $\rho_{k-1} = \rho_g$ by Lemma~\ref{pf:choice}. \ If
$\rho_{k-1} = \rho_g$ then $\rho_k = \rho_g$ by
Lemma~\ref{pf:goalabs}. \ Therefore, $\sigma_k$ occurs only if
$\rho_{k-1} = \sigma_{k-1}$. \ In this case the agent takes action $A^{i_k}$. \
If $A^{i_k}_j$ is applied to the system with $j < d+2$, $\rho_k$ is the goal state by
Lemma~\ref{pf:togoal}. \
Therefore, the system transitions to $\sigma_k$ exactly when $\rho_{k-1} =
\sigma_{k-1}$ and $A^{i_k}_{d+2}$ is applied. \ By induction
\begin{equation}
\sigma_{k-1} = \frac{K_{i_{k-1}}...K_{i_1}d(\rho_0) K_{i_1}^\dag
  ... K_{i_{k-1}}^\dag\oplus 0}{\Tr(K_{i_{k-1}}... K_{i_1}d(\rho_0)
  K_1^\dag ... K_{i_{k-1}}^\dag)}.
\end{equation}
Note that
\begin{equation}
d(\sigma_{k-1}) = \frac{K_{i_{k-1}}...K_{i_1}d(\rho_0) K_{i_1}^\dag
  ... K_{i_{k-1}}^\dag}{\Tr(K_{i_{k-1}}... K_{i_1}d(\rho_0)
  K_1^\dag ... K_{i_{k-1}}^\dag)}.
\label{eqn:sigk-1}
\end{equation}
Then
\begin{equation}
\sigma_k =
\frac{A^{i_k}_{d+2}\sigma_{k-1}A^{i_k}_{d+2}}{\Tr(A^{i_k}_{d+2}\sigma_{k-1}
 {A^{i_k}_{d+2}}^\dag)} =
\frac{K_{i_k}d(\sigma_{k-1})K_{i_k}^\dag\oplus
  0}{\Tr(K_{i_k}d(\sigma_{k-1}) K_{i_k}^\dag)}
\label{eqn:sigk}
\end{equation}
using Lemma~\ref{pf:trace}. \ Using Equation~\ref{eqn:sigk-1} for
$d(\sigma_{k-1})$, we have
\begin{equation}
K_{i_k}d(\sigma_{k-1})K^\dag_{i_k}
= \frac{K_{i_k}...K_{i_1}d(\rho_0)K_{i_1}^\dag... K_{i_k}^\dag}
{\Tr\left(K_{i_{k-1}}...K_{i_1}d(\rho_0)K_1
  ... K_{i_{k-1}}\right)},
\label{eqn:dsigk1}
\end{equation}
and
\begin{align}
\begin{split}
&\Tr(K_{i_k}d(\sigma_{k-1})K^\dag_{i_k})\\
&= \Tr\left(\frac{K_{i_k}...K_{i_1}d(\rho_0)K_{i_1}^\dag... K_{i_k}^\dag}
       {\Tr\left(K_{i_{k-1}}...K_{i_1}d(\rho_0)K_1
         ... K_{i_{k-1}}\right)}\right)
\end{split}\\
&= \frac{\Tr\left(K_{i_k}...K_{i_1}d(\rho_0)K_{i_1}^\dag... K_{i_k}^\dag\right)}{\Tr\left(K_{i_{k-1}}...K_{i_1}d(\rho_0)K_1
         ... K_{i_{k-1}}\right)},
\label{eqn:dsigk2}
\end{align}
Substituting equations~\ref{eqn:dsigk1} and \ref{eqn:dsigk2} for the
numerator and denominator of
equation~\ref{eqn:sigk} respectively, and canceling the traces, we find
\begin{equation}
\sigma_k = \frac{K_{i_k} ... K_{i_1}d(\rho_0) K_{i_1}... K_{i_k}\oplus
       0}{\Tr(K_{i_k} ... K_{i_1}d(\rho_0)
       K_{i_1}^\dag... K_{i_k}^\dag)}.
\end{equation}
\end{ProofEnv}

Now that we know how the state evolves, we can show that the
probability that the system is not in the goal state after taking actions
$\{A^{i_1}, ..., A^{i_n}\}$ should correspond to the
probability of observing measurements $\{i_1, ..., i_n\}$ in the
original QMOP instance.

\begin{ProofEnv}{Lemma}{}
\label{pf:goalprob}
Let ${\bf S} = \left\{K_1, ..., K_{\cal K}\right\}$ with Kraus
matrices of dimension $d$ be the superoperator from a QMOP instance and let
$Q({\bf S}) = \left\langle S, \Omega, {\cal A}, \rho_0,
\rho_g\right\rangle$ be the
corresponding goal QOMDP. \ Let $\pi$ be any policy and $\{\sigma_k\}$
be the policy path for $\pi$. \ Assume $\pi(\sigma_{j-1}) =
A^{i_j}$. \ The probability that
$\rho_n$ is not $\rho_g$ is
\begin{equation}
\Pr\left(\rho_n \neq \rho_g\right) =
\Tr(K_{i_n}...K_{i_1}d(\rho_0)K_{i_1}^\dag
... K_{i_n}^\dag).
\end{equation}

{\bf Proof:} First consider the probability that $\rho_n$ is not
$\rho_g$ given that $\rho_{n-1} \neq \rho_g$. \ By
Lemma~\ref{pf:choice}, if $\rho_{n-1} \neq \rho_g$ then $\rho_{n-1} =
\sigma_{n-1}$. \ By Lemma~\ref{pf:evol},
\begin{equation}
\sigma_{n-1} = \frac{K_{i_{n-1}}...K_{i_1}d(\rho_0)K_{i_1}^\dag
  ... K_{i_{n-1}}^\dag\oplus 0}{\Tr(K_{i_{n-1}}...K_{i_1}d(\rho_0)
  K_{i_1}^\dag ... K_{i_{n-1}}^\dag)}
\end{equation}
so
\begin{equation}
d(\sigma_{n-1}) = \frac{K_{i_{n-1}}...K_{i_1}d(\rho_0)K_{i_1}^\dag
  ... K_{i_{n-1}}^\dag}{\Tr(K_{i_{n-1}}...K_{i_1}d(\rho_0)
  K_{i_1}^\dag ... K_{i_{n-1}}^\dag)}.
\end{equation}
If $A^{i_n}_j$ for $j < d+2$ is applied then $\rho_n$ will be
$\rho_g$. \ Thus the probability that $\rho_n$ is {\em not} $\rho_g$ is
the probability that $A^{i_n}_{d+2}$ is applied:
\begin{align}
\begin{split}
\Pr&(\rho_n \neq \rho_g | \rho_{n-1}\neq
\rho_g )\\
&= \Tr(A^{i_n}_{d+2}\sigma_{n-1}{A^{i_n}_{d+2}}^\dag)
\end{split}\\
&= \Tr(K_{i_n}d(\sigma_{n-1})K^\dag_{i_n}\oplus 0)\\
&= \Tr(K_{i_n}d(\sigma_{n-1})K^\dag_{i_n})\\
&= \frac{\Tr(K_{i_n}...K_{i_1}d(\rho_0) K_{i_1}^\dag
  ... K_{i_n}^\dag)}{\Tr(K_{i_{n-1}}...K_{i_1}d(\rho_0)
  K_{i_1}^\dag ... K_{i_{i-1}}^\dag)}.
\end{align}
Note that $\Pr(\rho_n \neq \rho_g | \rho_{n-1} =
\rho_g) = 0$ by Lemma~\ref{pf:goalabs}. \ The total
probability that $\rho_n$ is not $\rho_g$ is
\begin{align*}
&\Pr(\rho_n \neq \rho_g)\\[0.1in]
&=\Pr(\rho_n \neq \rho_g \cap
\rho_{n-1} \neq \rho_g) + \Pr(\rho_n \neq \rho_g \cap \rho_{n-1}
= \rho_g)\\[0.1in]
&=\begin{aligned}
&\Pr(\rho_n\neq \rho_g | \rho_{n-1} \neq
\rho_g)\Pr(\rho_{n-1} \neq \rho_g) \ +\\
&\Pr(\rho_n\neq \rho_g |
\rho_{n-1} = \rho_g)\Pr(\rho_{n-1} = \rho_g)
\end{aligned}\\[0.1in]
&=\begin{aligned}
&\Pr(\rho_n \neq \rho_g | \rho_{n-1} \neq
\rho_g)\Pr(\rho_{n-1} \neq \rho_g | \rho_{n-2} \neq
\rho_g)\\
&...\Pr(\rho_1 \neq \rho_g | \rho_0 \neq
\rho_g)
\end{aligned}\\[0.1in]
&= \prod_{k=1}^n \frac{\Tr(K_{i_k}...K_{i_1}d(\rho_0) K_{i_1}^\dag
  ... K_{i_k}^\dag)}{\Tr(K_{i_{k-1}}...K_{i_1}d(\rho_0)
  K_{i_1}^\dag ... K_{i_{k-1}}^\dag)}\\[0.1in]
&= \Tr(K_{i_n}...K_{i_1}d(\rho_0) K_{i_1}^\dag
... K_{i_n}^\dag).
\end{align*}
\end{ProofEnv}

Since the probability that the agent observes the sequence of measurements
$\{i_1, ..., i_n\}$ is the same as the probability that the sequence
of actions $\{A^{i_1}, ..., A^{i_n}\}$ does not reach the goal state,
we can solve QMOP by solving
an instance of goal-state reachability for a QOMDP. \ Since QMOP is known to be
undecidable, this proves that goal-state reachability is also
undecidable for QOMDPs.

\begin{ProofEnv}{Theorem}{ (Undecidability of Goal-State Reachability
    for QOMDPs)}
\label{pf:qgrp}
The goal-state reachability problem for QOMDPs is undecidable.

{\bf Proof:}  As noted above, it suffices to show that we can reduce the quantum measurement occurrence
problem (QMOP) to goal-state reachability for QOMDPs.

Let ${\bf S} = \left\{K_1, ..., K_{\cal K}\right\}$ be the
superoperator from an instance of
QMOP with Kraus matrices of dimension $d$ and
let $Q({\bf S}) = \left\langle S, \Omega, {\cal A}, \rho_0, \rho_g\right\rangle$
be the corresponding goal QOMDP. \ By Theorem~\ref{pf:qomdp}, $Q({\bf
  S})$ is a goal QOMDP. \ We show that there is a policy that can reach
$\rho_g$ from $\rho_0$ with probability $1$ in a finite number of steps
if and only if there is some
finite sequence $\{i_1, ..., i_n\}$ such that $K_{i_1}^\dag
... K_{i_n}^\dag K_{i_n}...K_{i_1} = 0$.

First assume there is some sequence $\{i_1, ..., i_n\}$ such that
$K_{i_1}^\dag... K_{i_n}^\dag K_{i_n}...K_{i_1} = 0$. \ Consider the
time-dependent policy that takes action $A^{i_k}$ in after $k$ time
steps no matter the state. \
By
Lemma~\ref{pf:goalprob}, the probability that
this policy is not in the goal state after $n$ time steps is
\begin{align}
\Pr(\rho_n \neq \rho_g) &=
\Tr(K_{i_n}... K_{i_1}d(\rho_0)K_{i_1}^\dag...K_{i_n}^\dag)\\
&= \Tr(K_{i_1}^\dag... K_{i_n}^\dag K_{i_n} ... K_{i_1}d(\rho_0))\\
&= \Tr(0)\\
&= 0
\end{align}
using that $\Tr(AB) = \Tr(BA)$ for all matrices $A$ and $B$.
Therefore this policy reaches the goal state with probability $1$ after
$n$ time steps. \ As we have said, time cannot help goal decision
processes since nothing changes with time. \ Therefore, there is also a
purely state-dependent policy (namely the one that assigns $A^{i_k}$ to
$\sigma_k$ where $\sigma_k$ is the $k^{\text{th}}$ state reached when following
$\pi$) that can reach the goal state with probability $1$.

Now assume there is some policy $\pi$ that reaches the goal state with
probability $1$ after $n$ time steps. \ Let $\{\sigma_k\}$ be the policy
path and assume $\pi(\sigma_{k-1}) = A^{i_k}$. \ By
Lemma~\ref{pf:goalprob}, the probability that the state at time step
$n$ is not $\rho_g$ is
\begin{align}
\Pr(\rho_n \neq \rho_g | \pi) &=
\Tr(K_{i_1}...K_{i_n}d(\rho_0)K_{i_1}^\dag... K_{i_n}^\dag)\\
&=\Tr(K_{i_1}^\dag... K_{i_n}^\dag K_{i_n}...K_{i_i}d(\rho_0)).
\end{align}
Since $\pi$ reaches the goal state with probability $1$ after $n$
time steps, we must have that the above quantity is $0$. \
By construction $d(\rho_0)$ is full rank, so for the trace to
be $0$ we must have
\begin{equation}
K_{i_1}^\dag... K_{i_n}^\dag K_{i_n}...K_{i_i} = 0.
\end{equation}

Thus we can reduce the quantum measurement occurrence problem to
the goal-state reachability problem for QOMDPs, and the goal-state reachability
problem is undecidable for QOMDPs.
\end{ProofEnv}

\subsection{Decidability of Goal-State Reachability for POMDPs}

The goal-state reachability problem for POMDPs is decidable. \ This is a
known result~\cite{rintanen04}, but we
reproduce the proof here, because it is interesting to see the
differences between classical and quantum probability that lead to decidability for
the former.

At a high level, the goal-state reachability problem is decidable for
POMDPs because stochastic transition matrices have strictly nonnegative
elements. \ Since we are interested in a probability $1$ event, we can
treat probabilities as binary: either
positive or $0$. \ This gives us a belief space with
$2^{|S|}$ states rather than a continuous one, and we can show that the
goal-state reachability problem is decidable for finite state spaces.

\begin{DefnEnv}{Binary Probability MDP}
Given a goal POMDP $P = \langle S, A, \Omega, T, O, \vec{b}_0, g\rangle$, let
$M(P) = \langle B, A, \tau, \vec{b}_0, \vec{b}_g\rangle$ be the
corresponding goal belief
MDP with $\tau^{ao}$ defined according to
Equation~\ref{eqn:transmat}. \ Throughout this section, we assume without loss of generality that $g$
is the $|S|^{\text{th}}$ state in $P$ so $\left(\vec{b}_g\right)_i = \delta_{i|S|}$. \
The {\em binary probability MDP} is
an MDP $D(P) = \langle \Ze_{\{0,1\}}^{|S|}, A, Z, \vec{z}_0, \vec{z}_g\rangle$
where $\left(\vec{z}_g\right)_i = \delta_{i|S|}$ and  $\left(\vec{z}_0\right)_i
= 1$ if and only if $(\vec{b}_0)_i > 0$. \ The transition
function $Z$ for action $a$ non-deterministically applies the function
$Z^{ao}$ to $\vec{z}$. \ For $\vec{z}\in
\Ze_{\{0,1\}}^{|S|}$, the result of $Z^{ao}$ acting on $\vec{z}$ is
\begin{equation}
Z^{ao}(\vec{z})_i = \left\{\begin{array}{ll}
1 & \mbox{if } \left(\tau^{ao}\vec{z}\right)_i > 0\\
0 & \mbox{if } \left(\tau^{ao}\vec{z}\right)_i = 0.
\end{array}\right.
\label{eqn:zao}
\end{equation}
Let
\begin{equation}
P^o_a(\vec{z}) = \left\{\begin{array}{ll}
1 & \mbox{if } \tau^{ao}\vec{z} \neq \vec{0}\\
0 & \mbox{else}.
\end{array}\right.
\end{equation}
If action $a$ is taken in state $\vec{z}$, $Z^{ao}$ is applied with probability
\begin{equation}
\Pr\left(Z^{ao}|a, \vec{z}\right) = \left\{\begin{array}{ll}
\frac{1}{\sum_{o'\in \Omega}P^{o'}_a(\vec{z})} & \mbox{if }
P^a_o(\vec{z}) > 0\\
0 & \mbox{else.}
\end{array}\right.
\end{equation}
Note that the vector of all zeros is unreachable, so the state space is
really of size $2^{|S|}-1$.
\end{DefnEnv}

We first show that we can keep track of whether each entry in the
belief state is zero or not
just using the binary probability MDP. \ This lemma uses the fact that
classical probability involves nonnegative numbers only.

\begin{ProofEnv}{Lemma}{}
\label{pf:bprop}
Let $P = \langle S, A, \Omega, T, O, \vec{b}_0, g\rangle$
be a goal-state POMDP
and let $D(P) = \langle \Ze_{\{0,1\}}^{|S|}, A, Z, \vec{z}_0,
\vec{z}_g\rangle$ be the
associated binary probability MDP. \ Assume we have $\vec{z}$ and
$\vec{b}$ where $\vec{z}_i = 0$ if and only if $\vec{b}_i = 0$. \ Let
$$\vec{z}^{ao} =  Z^{ao}(\vec{z})$$
and
$$\vec{b}^{ao} =
\frac{\tau^{ao}\vec{b}}{\left|\tau^{ao}\vec{b}\right|_1}.$$
Then $\vec{z}^{ao}_i = 0$ if and only if $\vec{b}^{ao}_i = 0$. \
Moreover, $P^{o}_a(\vec{z}) = 0$ if and only if
$\left|\tau^{ao}\vec{b}\right|_1 = 0$.

{\bf Proof:}  Using the definition of $Z^{ao}$ from Equation~\ref{eqn:zao},
\begin{equation}
\vec{z}^{ao}_i = Z^{ao}(\vec{z})_i = \left\{\begin{array}{ll}
1 &\mbox{if } \left(\tau^{ao}\vec{z}\right)_i > 0\\
0 &\mbox{else}.
\end{array}\right.
\end{equation}
Let $N = \left|\tau^{ao}\vec{b}\right|_1$. \ Then
\begin{equation}
\label{eqn:bup}
\vec{b}^{ao}_i =
\frac{1}{N}
\sum_{j=1}^{|S|}\tau^{ao}_{ij}\vec{b}_j.
\end{equation}
Firstly assume $\vec{b}^{ao}_i = 0$. \ Since $\tau^{ao}_{ij} \geq 0$
and $\vec{b}_j\geq 0$, we must have that every term in the sum
in Equation~\ref{eqn:bup} is $0$ individually\footnote{This holds
  because probabilities are nonnegative. \ A similar
  analysis in the quantum case would fail at this step.}. \
Therefore, for all $j$, either $\tau^{ao}_{ij} = 0$ or
$\vec{b}_j = 0$. \ If $\vec{b}_j = 0$ then $\vec{z}_j = 0$
so $\tau^{ao}_{ij}\vec{z}_j = 0$. \ If
  $\tau^{ao}_{ij} = 0$ then clearly
  $\tau^{ao}_{ij}\vec{z}_j = 0$. \ Therefore
\begin{equation}
0 = \sum_{j=1}^{|S|}\tau^{ao}_{ij}\vec{z}_j =
    \left(\tau^{ao}\vec{z}\right)_i = \vec{z}^{ao}_i.
\end{equation}

Now assume $\vec{b}^{ao}_i > 0$. \ Then there must be at least one term in the
sum in Equation~\ref{eqn:bup} with $\tau^{ao}_{ik}\vec{b}_k
> 0$. \ In this case, we must have both $\tau^{ao}_{ik} >
0$ and $\vec{b}_k > 0$. \ If $\vec{b}_k > 0$ then
$\vec{z}_k > 0$. \ Therefore
\begin{equation}
\vec{z}^{ao}_i = \left(\tau^{ao}\vec{z}\right)_i =
\sum_{j=1}^{|S|}\tau^{ao}_{ij}\vec{z}_j =
\sum_{j\neq k}\tau^{ao}_{ij}\vec{z}_j +
\tau^{ao}_{ik}\vec{z}_k > 0.
\end{equation}
Since $\vec{b}^{ao}_i \geq 0$ and $\vec{z}^{ao}_i > 0$, we have shown
that $\vec{z}^{ao}_i = 0$ exactly when $\vec{b}^{ao}_i = 0$.

Now assume $\left|\tau^{ao}\vec{b}\right|_1 = 0$. \ This is true only if
$\tau^{ao}_{ij}\vec{b}_j = 0$ for all $i$ and $j$. \ Thus by the same reasoning
as above
$\tau^{ao}_{ij}\vec{z}_j = 0$ for all $i$ and $j$ so $\tau^{ao}\vec{z} = \vec{0}$
and $P^o_a(\vec{z}) = 0$.

Now let $\left|\tau^{ao}\vec{b}\right|_1 > 0$. \ Then there is some $k$
with $\tau^{ao}_{ik}\vec{z}_k > 0$ by the same reasoning as above. \
Therefore $\tau^{ao}\vec{z} \neq \vec{0}$ so $P^o_a(\vec{z}) = 1$.
\end{ProofEnv}

We now show that the agent can reach the goal in the binary probability MDP
with probability $1$ if and only if it could reach the goal in the
original POMDP with probability $1$. \ We
do each direction in a separate lemma.

\begin{ProofEnv}{Lemma}{}
\label{pf:pgrpf}
Let $P = \langle S, A, \Omega, T, O, \vec{b}_0, g\rangle$ be a
goal POMDP and let $D(P) =
\langle \Ze_{\{0,1\}}^{|S|}, A, Z, \vec{z}_0, \vec{z}_g\rangle$ be the corresponding
binary probability MDP. \ If there is a policy $\pi^D$ that reaches the goal with
probability $1$ in a finite number of steps in $D(M)$ then there is a
policy that reaches the goal in a finite number of steps with
probability $1$ in the belief MDP $M(P) = \left\langle B, A, \tau,
\vec{b}_0, \vec{b}_g\right\rangle$.

{\bf Proof:}  For
$\vec{b} \in B$ define $z(\vec{b})$ to be the single state $\vec{z}\in
\Ze^n_{\{0,1\}}$ with $\vec{z}_i = 0$ if and only if $\vec{b}_i = 0$. \ Let
$\pi$ be the policy for $M(P)$ with $\pi(\vec{b}) =
\pi^D(z(\vec{b}))$. \ Let $\vec{b}^0, \vec{b}^1, ..., \vec{b}^n$ be
some sequence of beliefs of length $n+1$ that can be created by following policy
$\pi$ with observations $\{o_{i_1}, ..., o_{i_n}\}$. \ Then
\begin{equation}
\vec{b}^{k+1} =
\frac{\tau^{\pi(\vec{b}^k)o_{i_k}}\vec{b}^k}{\left|\tau^{\pi(\vec{b}^k)o_{i_k}}\vec{b}^k\right|_1}
=
\frac{\tau^{\pi^D(z(\vec{b}^k))o_{i_k}}\vec{b}^k}{\left|\tau^{\pi^D(z(\vec{b}^k))
    o_{i_k}}\vec{b}^k\right|_1}.
\end{equation}
Define $a_k = \pi^D(z(\vec{b}^k))$.
Consider the set of states $\vec{z}^0, \vec{z}^1, ..., \vec{z}^n$ with
$\vec{z}^{k+1} = Z^{\pi^D(\vec{z}^k)o_{i_k}}\left(\vec{z}^k\right).$ \
We show by induction
that $\vec{z}^k = z(\vec{b}^k)$.

{\em Base Case $(k = 0)$}: We have $\vec{z}^0 = z(\vec{b}^0)$ by
definition.

{\em Induction Step}: Assume that $\vec{z}^k = z(\vec{b}^k)$. \
Then
\begin{equation}
\vec{z}^{k+1} = Z^{\pi^D(\vec{z}^k)o_{i_k}}(\vec{z}^k) =
Z^{\pi^D(z(\vec{b}^k))o_{i_k}}(\vec{z}^k) = Z^{a_ko_{i_k}}(\vec{z}^k)
\end{equation}
by induction. \ Now
\begin{equation}
\vec{b}^{k+1} =
\frac{\tau^{a_{i_k}o_{i_k}}\vec{b}^k}{\left|\tau^{a_{i_k}o_{i_k}}\vec{b}^k\right|_1}.
\end{equation}
Therefore $\vec{z}^{k+1} = z(\vec{b}^{k+1})$ by Lemma~\ref{pf:bprop}.

We must also show that the sequence $\vec{z}^0, \vec{z}^1, ...,
\vec{z}^n$ has nonzero probability of occurring while following
$\pi^D$. \ We must have that $P^{o_{i_k}}_{a_k} > 0$
for all $k$. \ We know that $\vec{b}^0, \vec{b}^1, ..., \vec{b}^n$ can
be created by
following $\pi$ so the probability of $\vec{b}^0, \vec{b}^1, ...,
\vec{b}^n$ is greater
than $0$. \ Therefore, we must have
\begin{equation}
\Pr(o | a_k, \vec{b}^k) =
\left|\tau^{a_ko_{i_k}}\vec{b}^k\right|_1 > 0
\end{equation}
for all $k$, so Lemma~\ref{pf:bprop} gives us that $P^{o_{i_k}}_{a_k} >
0$ for all $k$. \ Thus $\{\vec{z}^0,
..., \vec{z}^n\}$ is a possible sequence of states seen while
following policy $\pi^D$ in the MDP $D(P)$. \ Since $\pi^D$
reaches the goal state with probability $1$ after $n$ time steps, we have
$\vec{z}^n = \vec{z}_g$. \ Therefore, since $\vec{z}^n =
z(\vec{b}^n)$, we must have $\vec{b}^n_i = 0$ for all $i\neq |S|$, and
only $\vec{b}^n_{|S|} > 0$. \ Since $|\vec{b}^n|_1 = 1$, we have
$\vec{b}^n_{|S|} = 1$. \ Thus $\vec{b}^n = \vec{b}_g$ and $\pi$ also
reaches the goal state with nonzero probability after $n$ time steps.

\end{ProofEnv}

\begin{figure}
\centering
\includegraphics[width=\columnwidth]{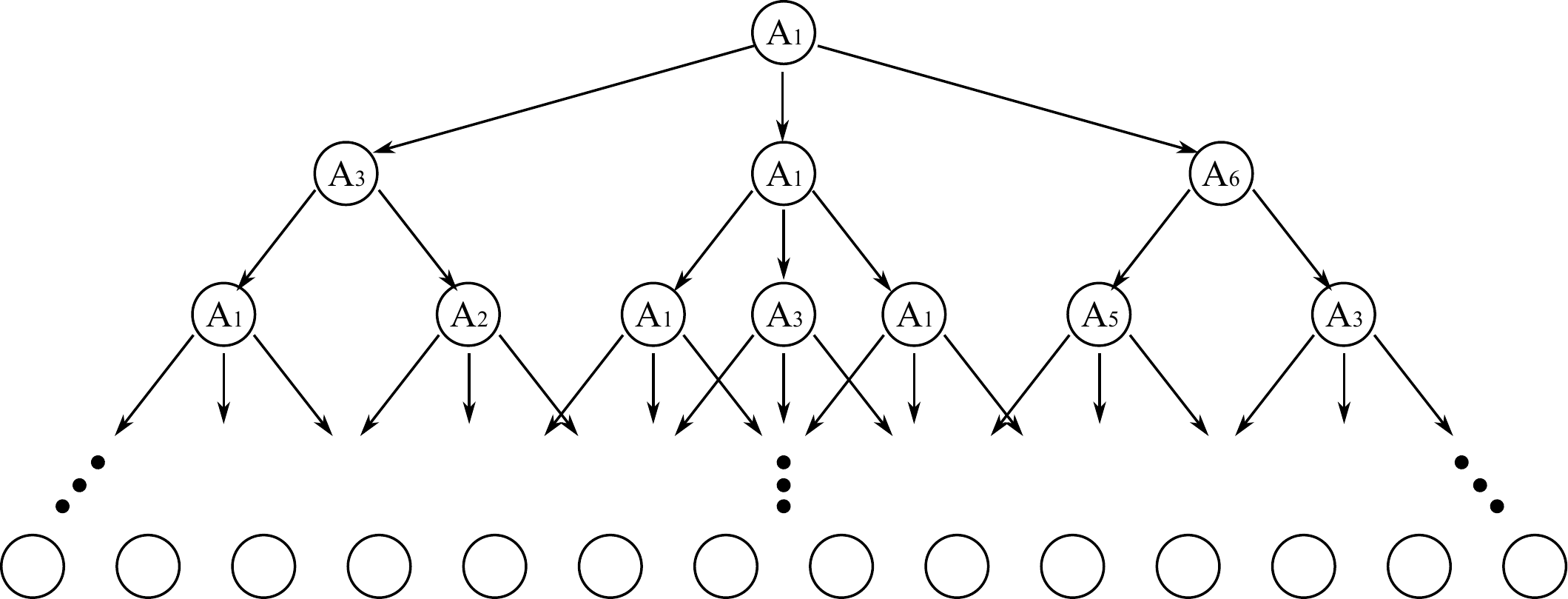}
\caption{A policy in an MDP creates a tree. \ Here, the agent takes action $A_1$ in the
  starting state, which can transitions the world state nondeterministically to
  three other possible states. \ The policy specifies an action of
  $A_3$ for the state on the left, $A_1$ for the state in the middle
  and $A_6$ for the state on the right. \ Taking these actions
  transition these states nondeterministically. \ This tree
  eventually encapsulates all states that can be reached with nonzero
  probability from the starting state under a particular policy. \ The
  goal can be reached with probability $1$ if there is some depth below
  which every node is the goal state.}
\label{fig:mdptree}
\end{figure}

\begin{ProofEnv}{Lemma}{}
\label{pf:pgrpb}
Let $P = \langle S, A, \Omega, T, O, \vec{b}_0, g\rangle$ be a
goal POMDP and let $D(P) =
\langle \Ze_{\{0,1\}}^{|S|}, A, Z, \vec{z}_0, \vec{z}_g\rangle$ be the corresponding
binary probability MDP. \ If there is a policy $\pi$ that reaches the goal with
probability 1 in a finite number of steps in the belief state MDP
$B(M) = \langle B, A, \tau, \vec{b}_0, \vec{b}_g\rangle$
then there is a
policy that reaches the goal in a finite number of steps with
probability $1$ in $D(P)$.

{\bf Proof:} MDP policies create trees of states and action choices as
shown in Figure~\ref{fig:mdptree}. \ Consider the tree $\pi_T$ formed
by $\pi$. \ Nodes at depth $n$ or
greater are guaranteed to be
$\vec{b}_g$. \ For $\vec{z}\in \Ze^{|S|}_{\{0,1\}}$, we let $b(\vec{z})$
be the deepest state in $\pi_T$ for which $\vec{b}_i = 0$ if and only
if $\vec{z}_i = 0$. \ If there are multiple states for which this is
true at the same level, we choose the leftmost one. \
If no such state is
found in $\pi_T$, we set $b(\vec{z}) = \vec{b}_g$. \ We define a policy
$\pi^D$ for $D(P)$ by $\pi^D(\vec{z}) = \pi(b(\vec{z}))$. \ Let
$\vec{z}^0, \vec{z}^1, ..., \vec{z}^n$ be any sequence of states that
can be created
by following policy $\pi^D$ in $D(P)$ for $n$ time steps. \ Define $a_k =
\pi^D(\vec{z}^k)$ and define $i_k$ as the smallest number such that
$\vec{z}^{k+1} = Z^{a_ko_{i_k}}(\vec{z}^k)$ (some such
$Z^{a_ko_{i_k}}$ exists since $\vec{z}^0, ..., \vec{z}^n$ can be
created by following $\pi^D$). \ Now consider
$b(\vec{z}^k)$. \ We show by induction that this state is at least at
level $k$ of $\pi_T$.

{\em Base Case $(k = 0)$}: We know that $\vec{b}^0_i = 0$ if and only if
$\vec{z}^0_i = 0$ so $b(\vec{z}^0)$ is at least at level 0 of $\pi_T$.

{\em Induction Step}: Assume that $\vec{z}^k$ is at least at level $k$
of $\pi_T$. \ Then
\begin{equation}
\vec{z}^{k+1} = Z^{a_ko_{i_k}}\left(\vec{z}^k\right).
\end{equation}
Therefore by Lemma~\ref{pf:bprop},
\begin{equation}
\vec{b}' =
\frac{\tau^{a_ko_{i_k}}b(\vec{z}^k)}{\left|\tau^{a_ko_{i_k}}b(\vec{z}^k)\right|_1}
\end{equation}
has entry $i$ 0 if and only if $\vec{z}^{k+1}_i = 0$. \ Now
$P^{a_k}_{o_k}(\vec{z}^k) \neq 0$ only if
$|\tau^{a_ko_{i_k}}b(\vec{z}^k)|_1 \neq 0$ also by
Lemma~\ref{pf:bprop}. \ Since $\vec{z}^1, ..., \vec{z}^n$ is a branch
of $\pi^D$, we must have $P^{a_k}_{o_k} > 0$. \ Therefore
$|\tau^{a_ko_{i_k}}b(\vec{z}^k)|_1 > 0$. \ Now $a_k =
\pi(b(\vec{z}^k))$ so $\vec{b}'$ is a child
of $b(\vec{z}^k)$ in $\pi_T$. \ Since, by induction, the level of
$b(\vec{z}^k)$ is at least $k$, the level of $\vec{b}'$ is at least
$k+1$. \ Now $\vec{b} = b(\vec{z}^{k+1})$ is the deepest state in the
tree with $\vec{b}_i = 0$ if and only if $\vec{z}^{k+1}_i = 0$ so
level of $b(\vec{z}^{k+1})$ is at least the level of $\vec{b}'$. \
Therefore $b(\vec{z}^{k+1})$ has level at least $k+1$.

Thus the level of $b(\vec{z}^n)$ is at least $n$. \ We have
$b(\vec{z}^n) = \vec{b}_g$ since $\pi$ reaches the goal state in at most $n$ steps. \ Since $b(\vec{z}^n)_i = \delta_{i|S|}$, we
have that $\vec{z}^n = \vec{z}_g$. \ Therefore $\pi^D$ is
a policy for $D(P)$ that reaches the goal with probability $1$ in at most $n$ steps.
\end{ProofEnv}

We have now reduced goal-state reachability for POMDPs to goal-state
reachability for finite-state MDPs. \ We briefly show that the latter is
decidable.

\begin{ProofEnv}{Theorem}{ (Decidability of Goal-State Reachability
    for POMDPs)}
The goal-state reachability problem for POMDPs is decidable.

{\bf Proof:} We showed in Lemmas~\ref{pf:pgrpf}
and~\ref{pf:pgrpb} that goal-state reachability for POMDPs can be
reduced to goal-state reachability for a finite state MDP. \ Therefore,
there are only $O(|A|^{|S|})$ possible policies (remember that for goal
decision processes, we need only consider time independent policies). \ Given
a policy $\pi$, we can evaluate it by creating a directed graph $G$ in which we
connect state $s_i$ to state $s_j$ if $\tau(s_i, \pi(s_i), s_j) >
0$. \ The policy $\pi$ reaches the goal from the starting state in a
finite number of steps with
probability $1$ if the goal is reachable from the starting state in $G$ and no
cycle is reachable. \ The number of nodes in the graph is at most the
number of states in the MDP so we can clearly decide this problem. \ Thus
goal-state reachability is decidable for POMDPs.
\end{ProofEnv}

\subsection{Other Computability Separations}
\label{sec:more}

Although we looked only at goal-state reachability here, we conjecture that there are other similar problems that are undecidable for QOMDPs despite being decidable for POMDPs.


For instance, the zero-reward policy problem is a likely candidate for
computability separation. \ In this problem, we still have a goal
QOMDP(POMDP) but states other than the goal state are allowed to have
zero reward. \ The problem is to decide whether the path to the goal
state is zero reward. \ This is known to be decidable for POMDPs, but
seems unlikely to be so for QOMDPs.

\section{Future Work}

We were only able to give an interesting
computability result for a problem about goal decision processes, which ignore the
reward function. \ It would be a great to prove a result about
QOMDPs that made nontrivial use of the reward function.

We also proved computability results, but did not consider algorithms for
solving any of the problems we posed beyond a very simple $\mathsf{PSPACE}$
algorithm for policy existence. \ Are there quantum analogues of POMDP
algorithms or even MDP ones?

\begin{acknowledgements}
This material is based upon work supported by the National
Science Foundation under Grants No. 0844626 and 1122374, as well as
an NSF Waterman Award.
\end{acknowledgements}

\bibliography{qomdps}

\begin{thebibliography}{10}%
\makeatletter
\providecommand \@ifxundefined [1]{%
 \ifx #1\undefined \expandafter \@firstoftwo
 \else \expandafter \@secondoftwo
\fi
}%
\providecommand \@ifnum [1]{%
 \ifnum #1\expandafter \@firstoftwo
 \else \expandafter \@secondoftwo
\fi
}%
\providecommand \enquote [1]{``#1''}%
\providecommand \bibnamefont  [1]{#1}%
\providecommand \bibfnamefont [1]{#1}%
\providecommand \citenamefont [1]{#1}%
\providecommand\href[0]{\@sanitize\@href}%
\providecommand\@href[1]{\endgroup\@@startlink{#1}\endgroup\@@href}%
\providecommand\@@href[1]{#1\@@endlink}%
\providecommand \@sanitize [0]{\begingroup\catcode`\&12\catcode`\#12\relax}%
\@ifxundefined \pdfoutput {\@firstoftwo}{%
 \@ifnum{\z@=\pdfoutput}{\@firstoftwo}{\@secondoftwo}%
}{%
 \providecommand\@@startlink[1]{\leavevmode\special{html:<a href="#1">}}%
 \providecommand\@@endlink[0]{\special{html:</a>}}%
}{%
 \providecommand\@@startlink[1]{%
  \leavevmode
  \pdfstartlink
   attr{/Border[0 0 1 ]/H/I/C[0 1 1]}%
   user{/Subtype/Link/A<</Type/Action/S/URI/URI(#1)>>}%
  \relax
 }%
 \providecommand\@@endlink[0]{\pdfendlink}%
}%
\providecommand \url  [0]{\begingroup\@sanitize \@url }%
\providecommand \@url [1]{\endgroup\@href {#1}{\urlprefix}}%
\providecommand \urlprefix [0]{URL }%
\providecommand \Eprint[0]{\href }%
\@ifxundefined \urlstyle {%
  \providecommand \doi [1]{doi:\discretionary{}{}{}#1}%
}{%
  \providecommand \doi [0]{doi:\discretionary{}{}{}\begingroup
  \urlstyle{rm}\Url }%
}%
\providecommand \doibase [0]{http://dx.doi.org/}%
\providecommand \Doi[1]{\href{\doibase#1}}%
\providecommand \bibAnnote [3]{%
  \BibitemShut{#1}%
  \begin{quotation}\noindent
    \textsc{Key:}\ #2\\\textsc{Annotation:}\ #3%
  \end{quotation}%
}%
\providecommand \bibAnnoteFile [2]{%
  \IfFileExists{#2}{\bibAnnote {#1} {#2} {\input{#2}}}{}%
}%
\providecommand \typeout [0]{\immediate \write \m@ne }%
\providecommand \selectlanguage [0]{\@gobble}%
\providecommand \bibinfo [0]{\@secondoftwo}%
\providecommand \bibfield [0]{\@secondoftwo}%
\providecommand \translation [1]{[#1]}%
\providecommand \BibitemOpen[0]{}%
\providecommand \bibitemStop [0]{}%
\providecommand \bibitemNoStop [0]{.\EOS\space}%
\providecommand \EOS [0]{\spacefactor3000\relax}%
\providecommand \BibitemShut [1]{\csname bibitem#1\endcsname}%
\bibitem{kaelbling98}%
  \BibitemOpen
  \bibfield{author}{%
  \bibinfo {author} {\bibfnamefont{L.~P.}\ \bibnamefont{Kaelbling}}, \bibinfo
  {author} {\bibfnamefont{M.~L.}\ \bibnamefont{Littman}},\ and\ \bibinfo
  {author} {\bibfnamefont{A.~R.}\ \bibnamefont{Cassandra}},\ }%
  \bibfield{journal}{%
  \bibinfo {journal} {Artificial Intelligence}\ }%
  \textbf{\bibinfo {volume} {101}},\ \bibinfo {pages} {99} (\bibinfo {year}
  {1998})%
  \bibAnnoteFile{NoStop}{kaelbling98}%
\bibitem{pineau03}%
  \BibitemOpen
  \bibfield{author}{%
  \bibinfo {author} {\bibfnamefont{J.}~\bibnamefont{Pineau}}, \bibinfo {author}
  {\bibfnamefont{G.}~\bibnamefont{Gordon}},\ and\ \bibinfo {author}
  {\bibfnamefont{S.}~\bibnamefont{Thrun}},\ }%
  in\ \emph{\bibinfo {booktitle} {Proceedings of the 18th International Joint
  Conference on Artificial Intelligence}},\ \bibinfo {editor} {edited by\
  \bibinfo {editor} {\bibfnamefont{G.}~\bibnamefont{Gottlob}}\ and\ \bibinfo
  {editor} {\bibfnamefont{T.}~\bibnamefont{Walsh}}}\ (\bibinfo {address}
  {Acapulco, Mexico},\ \bibinfo {year} {2003})\ pp.\ \bibinfo {pages}
  {1025--1032}%
  \bibAnnoteFile{NoStop}{pineau03}%
\bibitem{russell03}%
  \BibitemOpen
  \bibfield{author}{%
  \bibinfo {author} {\bibfnamefont{S.}~\bibnamefont{Russell}}\ and\ \bibinfo
  {author} {\bibfnamefont{P.}~\bibnamefont{Norvig}},\ }%
  \emph{\bibinfo {title} {{Artificial Intelligence: A Modern Approach}}},\
  \bibinfo {edition} {2nd}\ ed.\ (\bibinfo {publisher} {Pearson Hall},\
  \bibinfo {address} {New Jersey},\ \bibinfo {year} {2003})\ Chap.~\bibinfo
  {chapter} {17}, pp.\ \bibinfo {pages} {613--648}%
  \bibAnnoteFile{NoStop}{russell03}%
\bibitem{spaan05}%
  \BibitemOpen
  \bibfield{author}{%
  \bibinfo {author} {\bibfnamefont{M.~T.~J.}\ \bibnamefont{Spaan}}\ and\
  \bibinfo {author} {\bibfnamefont{N.}~\bibnamefont{Vlassis}},\ }%
  \bibfield{journal}{%
  \bibinfo {journal} {Journal of Artificial Intelligence Research}\ }%
  \textbf{\bibinfo {volume} {24}},\ \bibinfo {pages} {195} (\bibinfo {year}
  {2005})%
  \bibAnnoteFile{NoStop}{spaan05}%
\bibitem{smith04}%
  \BibitemOpen
  \bibfield{author}{%
  \bibinfo {author} {\bibfnamefont{T.}~\bibnamefont{Smith}}\ and\ \bibinfo
  {author} {\bibfnamefont{R.}~\bibnamefont{Simmons}},\ }%
  in\ \emph{\bibinfo {booktitle} {Proceedings of the 20th Conference on
  Uncertainty in Artificial Intelligence}}\ (\bibinfo {address} {Banff,
  Canada},\ \bibinfo {year} {2004})\ pp.\ \bibinfo {pages} {520--527}%
  \bibAnnoteFile{NoStop}{smith04}%
\bibitem{barry}%
  \BibitemOpen
  \bibfield{author}{%
  \bibinfo {author} {\bibfnamefont{J.}~\bibnamefont{Barry}}, \bibinfo {author}
  {\bibfnamefont{D.}~\bibnamefont{Barry}},\ and\ \bibinfo {author}
  {\bibfnamefont{S.}~\bibnamefont{Aaronson}},\ }%
  \enquote{\bibinfo {title} {{Quantum POMDPs}},}\ \bibinfo {howpublished}
  {{arXiv:1406.2858v1}} (\bibinfo {year} {2014})%
  \bibAnnoteFile{NoStop}{barry}%
\bibitem{ying}%
  \BibitemOpen
  \bibfield{author}{%
  \bibinfo {author} {\bibfnamefont{S.}~\bibnamefont{Ying}}\ and\ \bibinfo
  {author} {\bibfnamefont{M.}~\bibnamefont{Ying}},\ }%
  \enquote{\bibinfo {title} {{Reachability Analysis of Quantum Markov Decision
  Processes}},}\ \bibinfo {howpublished} {{arVix:1406.6146v2}} (\bibinfo {year}
  {2014})%
  \bibAnnoteFile{NoStop}{ying}%
\bibitem{ying01}%
  \BibitemOpen
  \bibfield{author}{%
  \bibinfo {author} {\bibfnamefont{S.}~\bibnamefont{Ying}}, \bibinfo {author}
  {\bibfnamefont{N.}~\bibnamefont{Yu}},\ and\ \bibinfo {author}
  {\bibfnamefont{M.}~\bibnamefont{Ying}},\ }%
  in\ \emph{\bibinfo {booktitle} {Proceedings of the 24th International
  Conference on Concurrency Theory}}\ (\bibinfo {address} {Buenos Aires,
  Argentina},\ \bibinfo {year} {2013})\ pp.\ \bibinfo {pages} {334--348}%
  \bibAnnoteFile{NoStop}{ying01}%
\bibitem{combes}%
  \BibitemOpen
  \bibfield{author}{%
  \bibinfo {author} {\bibfnamefont{J.}~\bibnamefont{Combes}}, \bibinfo {author}
  {\bibfnamefont{C.}~\bibnamefont{Ferrie}}, \bibinfo {author}
  {\bibfnamefont{C.}~\bibnamefont{Cesare}}, \bibinfo {author}
  {\bibfnamefont{M.}~\bibnamefont{Tiersch}}, \bibinfo {author}
  {\bibfnamefont{G.}~\bibnamefont{Milburn}}, \bibinfo {author}
  {\bibfnamefont{H.}~\bibnamefont{Briegel}},\ and\ \bibinfo {author}
  {\bibfnamefont{C.}~\bibnamefont{Caves}},\ }%
  \bibfield{journal}{%
  \bibinfo {journal} {arXiv},\ \bibinfo {pages} {1405.5656}}%
   (\bibinfo {year} {2014})%
  \bibAnnoteFile{NoStop}{combes}%
\bibitem{tiersch}%
  \BibitemOpen
  \bibfield{author}{%
  \bibinfo {author} {\bibfnamefont{M.}~\bibnamefont{Tiersch}}, \bibinfo
  {author} {\bibfnamefont{E.}~\bibnamefont{Ganahl}},\ and\ \bibinfo {author}
  {\bibfnamefont{H.}~\bibnamefont{Briegel}},\ }%
  \bibfield{journal}{%
  \bibinfo {journal} {arXiv},\ \bibinfo {pages} {1407.1535}}%
   (\bibinfo {year} {2014})%
  \bibAnnoteFile{NoStop}{tiersch}%
\bibitem{papadimitriou87}%
  \BibitemOpen
  \bibfield{author}{%
  \bibinfo {author} {\bibfnamefont{C.~H.}\ \bibnamefont{Papadimitriou}}\ and\
  \bibinfo {author} {\bibfnamefont{J.~N.}\ \bibnamefont{Tsitsiklis}},\ }%
  \bibfield{journal}{%
  \bibinfo {journal} {Mathematics of Operations Research}\ }%
  \textbf{\bibinfo {volume} {12}},\ \bibinfo {pages} {441} (\bibinfo {month}
  {August}\ \bibinfo {year} {1987})%
  \bibAnnoteFile{NoStop}{papadimitriou87}%
\bibitem{madani99}%
  \BibitemOpen
  \bibfield{author}{%
  \bibinfo {author} {\bibfnamefont{O.}~\bibnamefont{Madani}}, \bibinfo {author}
  {\bibfnamefont{S.}~\bibnamefont{Hanks}},\ and\ \bibinfo {author}
  {\bibfnamefont{A.}~\bibnamefont{Condon}},\ }%
  in\ \emph{\bibinfo {booktitle} {Proceedings of the 16th National Conference
  on Artificial Intelligence}}\ (\bibinfo {address} {Menlo Park, CA},\ \bibinfo
  {year} {1999})\ pp.\ \bibinfo {pages} {541--548}%
  \bibAnnoteFile{NoStop}{madani99}%
\bibitem{Note1}%
  \BibitemOpen
  \bibinfo {note} {Actually, the quantum operator acts on a product state of
  which the first dimension is $d$. \ In order to create quantum states of
  dimension $d$ probabilistically, the superoperator entangles the possible
  next states with a measurement register and then measures that register. \
  Thus the operator actually acts on the higher-dimensional product space, but
  for the purposes of this discussion, we can treat it as an operator that
  probabilistically maps states of dimension $d$ to states of dimension $d$.}%
  \bibAnnoteFile{Stop}{Note1}%
\bibitem{superoperator}%
  \BibitemOpen
  \bibfield{author}{%
  \bibinfo {author} {\bibfnamefont{M.}~\bibnamefont{Neilson}}\ and\ \bibinfo
  {author} {\bibfnamefont{I.}~\bibnamefont{Chuang}},\ }%
  \emph{\bibinfo {title} {Quantum Computation and Quantum Information}},\
  \bibinfo {edition} {10th}\ ed.\ (\bibinfo {publisher} {Cambridge University
  Press},\ \bibinfo {address} {Cambridge, UK},\ \bibinfo {year} {2011})%
  \bibAnnoteFile{NoStop}{superoperator}%
\bibitem{eisert12}%
  \BibitemOpen
  \bibfield{author}{%
  \bibinfo {author} {\bibfnamefont{J.}~\bibnamefont{Eisert}}, \bibinfo {author}
  {\bibfnamefont{M.~P.}\ \bibnamefont{Mueller}},\ and\ \bibinfo {author}
  {\bibfnamefont{C.}~\bibnamefont{Gogolin}},\ }%
  \bibfield{journal}{%
  \bibinfo {journal} {Physical Review Letters}\ }%
  \textbf{\bibinfo {volume} {108}} (\bibinfo {year} {2012})%
  \bibAnnoteFile{NoStop}{eisert12}%
\bibitem{rintanen04}%
  \BibitemOpen
  \bibfield{author}{%
  \bibinfo {author} {\bibfnamefont{J.}~\bibnamefont{Rintanen}},\ }%
  in\ \emph{\bibinfo {booktitle} {Proceedings of the 14th International
  Conference on Automated Planning and Scheduling}}\ (\bibinfo {publisher}
  {AAAI Press},\ \bibinfo {address} {Menlo Park, CA},\ \bibinfo {year} {2004})\
  pp.\ \bibinfo {pages} {345--354}%
  \bibAnnoteFile{NoStop}{rintanen04}%
\bibitem{Note2}%
  \BibitemOpen
  \bibinfo {note} {This holds because probabilities are nonnegative. \ A
  similar analysis in the quantum case would fail at this step.}%
  \bibAnnoteFile{Stop}{Note2}%
\end{thebibliography}%

\end{document}